\documentclass[manuscript,screen]{acmart}

\AtBeginDocument{%
  \providecommand\BibTeX{{%
    \normalfont B\kern-0.5em{\scshape i\kern-0.25em b}\kern-0.8em\TeX}}}

\setcopyright{acmcopyright}
\copyrightyear{2018}
\acmYear{2018}
\acmDOI{10.1145/1122445.1122456}

\acmJournal{JACM}
\acmVolume{37}
\acmNumber{4}
\acmArticle{111}
\acmMonth{8}

\usepackage{multirow}
\usepackage{graphicx}
\usepackage{subfigure}
\usepackage{float}
\usepackage[section]{placeins}

\begin{document}
%\begin{CJK*}{UTF8}{gbsn}

\title{A Static and Dynamic Attention Framework for Multi Turn Dialogue Generation}
\thanks{This manuscript is an extension of the authors' conference paper\cite{zhang2018context} which had published in COLING 2018. }

\author{Weinan Zhang}
\email{wnzhang@ir.hit.edu.cn}
\affiliation{%
  \institution{Research Center for Social Computing and Information Retrieval, Harbin Institute of Technology}
  \city{Harbin}
  \country{China}
}

\author{Yiming Cui}
\email{ymcui@iflytek.com}
\affiliation{%
 \institution{Research Center for Social Computing and Information Retrieval, Harbin Institute of Technology;}
  \city{Harbin}
  \institution{State Key Laboratory of Cognitive Intelligence, iFLYTEK Research}
  \city{Beijing}
  \country{China}
}

\author{Kaiyan Zhang}
\email{kyzhang@ir.hit.edu.cn}
\affiliation{%
 \institution{Research Center for Social Computing and Information Retrieval, Harbin Institute of Technology}
  \city{Harbin}
  \country{China}
  }

\author{Yifa Wang}
\email{yfwang@ir.hit.edu.cn}
\affiliation{%
 \institution{Research Center for Social Computing and Information Retrieval, Harbin Institute of Technology}
  \city{Harbin}
  \country{China}
  }

\author{Qingfu Zhu}
\email{qfzhu@ir.hit.edu.cn}
\affiliation{%
  \institution{Research Center for Social Computing and Information Retrieval, Harbin Institute of Technology}
  \city{Harbin}
  \country{China}
}

\author{Lingzhi Li}
\email{lzli@ir.hit.edu.cn}
\affiliation{%
 \institution{Research Center for Social Computing and Information Retrieval, Harbin Institute of Technology}
  \city{Harbin}
  \country{China}
}

\author{Ting Liu}
\authornote{Corresponding author.}
\email{tliu@ir.hit.edu.cn}
\affiliation{%
  \institution{Research Center for Social Computing and Information Retrieval, Harbin Institute of Technology}
  \city{Harbin}
  \state{Heilongjiang}
  \country{China}
}

\renewcommand{\shortauthors}{Weinan Zhang and Yiming Cui, et al.}

\begin{abstract}
Recently, research on open domain dialogue systems have attracted extensive interests of academic and industrial researchers. The goal of an open domain dialogue system is to imitate humans in conversations. Previous works on single turn conversation generation have greatly promoted the research of open domain dialogue systems. However, understanding multiple single turn conversations is not equal to the understanding of multi turn dialogue due to the coherent and context dependent properties of human dialogue.  Therefore, in open domain multi turn dialogue generation, it is essential to modeling the contextual semantics of the dialogue history, rather than only according to the last utterance. Previous research had verified the effectiveness of the hierarchical recurrent encoder-decoder framework on open domain multi turn dialogue generation. However, using RNN-based model to hierarchically encoding the utterances to obtain the representation of dialogue history still face the problem of a vanishing gradient. To address this issue, in this paper, we proposed a static and dynamic attention-based approach to model the dialogue history and then generate open domain multi turn dialogue responses. Experimental results on Ubuntu and Opensubtitles datasets verify the effectiveness of the proposed static and dynamic attention-based approach on automatic and human evaluation metrics in various experimental settings. Meanwhile, we also empirically verify the performance of combining the static and dynamic attentions on open domain multi turn dialogue generation. 
\end{abstract}





\maketitle

\section{Introduction}

Since the question ``\emph{Can machines think}?'' proposed by A.M. Turing~\cite{Turing1950} in 1950, passing the Turing test\footnote{\url{https://en.wikipedia.org/wiki/Turing_test}} becomes a long term goal for the artificial intelligence research.
As the Turing test is designed as an imitation game through human-machine conversation, the research of open domain conversations attracted wide attention of researchers.
However, until now, training an open domain conversational system to passing the Turing test is still a non-trivial task.
In the early work, an unsupervised clustering approach~\cite{ritter2010}, a phrase-based statistical machine translation approach~\cite{ritter2011} and a vector space model (VSM)-based approach~\cite{iris} are proposed to generate responses of open domain conversations.
Recently, with the blooming of deep learning techniques, especially the neural sequence-to-sequence learning (Seq2Seq) models, the generation of open domain conversational responses follows the framework of an end-to-end encoding and decoding process~\cite{s2snn,ncm,nrm,mrrnn,dpcm,persona,EMNLP17-1234,EMNLP17-1232}.
Previous works on short text conversation~\cite{nrm} (single turn conversation generation) successfully promote the research of generative approaches on open domain conversation generation.
However, a session of human conversations usually consists of multiple utterances so that it is often called a multi turn dialogue.
Therefore, in the imitation of human conversations, an open domain dialogue model should consider the historically utterances rather than the last utterance.
Table~\ref{example-intro} shows the impact of historically utterances(dialogue context) on response generation.
\begin{table}
\caption{An example of human-human conversation, where ``A'' and ``B'' denote two interlocutors.}
\label{example-intro}
\begin{tabular}{l|p{6cm}|lp{5cm}}
\multicolumn{2}{c}{\bf Conversation \#1}  & \multicolumn{2}{c}{\bf Conversation \#2} \\
\hline
A: & How about going to the Universal Studio together?  & A: & Hi, shall we go to the Starbucks? \\
B: & OK, let's go!  & B: & OK, let's go!  \\
A: & \emph{What do you like best there?}  & A: & \emph{What do you like best there?} \\
\hline
B: & \bf \emph{The Transformers is very exciting!}  & B: & \bf \emph{The cold brew.} \\
\hline
\end{tabular}
\end{table}

As shown in Table~\ref{example-intro}, when giving the same input message: ``\emph{What do you like best there?}'', different responses are separately generated by different speakers according to the different dialogue context.
However, an open domain conversation model, which does not consider the dialogue context, will generate a unique response according to the given message (the last utterance).
It will lead to the generation of incoherent human-computer dialogues and thus seriously impact the users' experience.
Therefore, recent works begin to model the coherence and context-aware generation~\cite{hred,vhred,ACL17-2036,ms20171,drl,shen2019modeling,zhang2019recosa,li2019dense,lan2019talk,zhou2019unsupervised} of open domain conversations.
An early context-sensitive dialogue generation model is the hierarchical recurrent encoder-decoder (HRED)~\cite{hred}.
Two RNNs are utilized to model each utterance in dialogue context and each word in an utterance in a unique framework.
Due to the generation of generic responses, they~\cite{vhred} further proposed a variational model based on the HRED model, which is called variational HRED (vHRED).
A stochastic latent variable is added at each utterance to improve the diversity of response generation.
Furthermore, the CVAE model~\cite{ACL17-1061} is proposed by introducing a conditional variational autoencoder into the encoder-decoder framework to learn the diversity of dialogue context for conversation generation.

Besides the introducing of stochastic latent variable to improve the diversity of generated responses, the attention mechanism is also utilized to model the dialogue context and generate responses.
Two typical research works are the HRAN~\cite{ms20171} and the WSI~\cite{ACL17-2036}.
The former is proposed to hierarchically model the interactions between words and utterances.
While, the latter models the dialogue context through a recurrent neural network in encoding step.

Previous research had verified the effectiveness of the hierarchical recurrent encoder-decoder framework on open-domain multi turn dialogue generation. 
However, using RNN-based model to hierarchically encoding the utterances to obtain the representation of dialogue context still face the problem of a vanishing gradient. 
Recently, ReCoSa~\cite{zhang2019recosa} is proposed to utilize self-attention to model the representations of dialogue context and utterance and their interactions.
From these works, we can draw the following conclusions:
\begin{itemize}
  \item First, the use of GRU~\cite{ms20171,ACL17-2036} or LSTM~\cite{zhang2019recosa} is verified to be effective on obtaining utterance representation.
  \item Second, the attention (or self-attention)-based approach~\cite{ms20171,zhang2019recosa} is more effective to model the contextual semantics of dialogue than the sequential integration approach~\cite{ACL17-2036}.
  \item Third, the representation of dialogue context can be improved by the interaction of self-attention in an utterance and the global attention between utterances and response~\cite{zhang2019recosa}.
\end{itemize}

Drawing the above advantages, in this paper, we proposed a static and dynamic attention framework for the generation of multi turn conversations. 
In the proposed framework, we present the way of integrating the global attention (static attention) and the local attention (dynamic attention) to model the multi turn dialogue and generate coherent and diverse conversational responses.
Furthermore, the proposed static and dynamic attention-based model is compatible to different granularities of input information, such as token-level and utterance-level information, different types of attentions, such as multi head attention and self-attention, and pretrained language model, such as GPT2, etc. 
Meanwhile, the proposed static and dynamic attentions can be integrated in various measures. 

\section{Related Work}
In the early work, a data-driven statistical machine translation (SMT)-based approach~\cite{ritter2011} had been proposed to generate responses in social media platform.
It verified that the SMT-based generation model outperforms the IR-based model in response generation.
It is also a representative work on open domain dialogue generation. 

Recently, the neural encoder-decoder framework for sequential generation are widely used to dialogue generation~\cite{s2snn,nrm,mrrnn,zhu2016learning,EMNLP17-1234,EMNLP17-1232,hred,vhred,zhang2019recosa,lu2020improving,ling2020leveraging}.
The first sequence-to-sequence (Seq2Seq)-based neural network~\cite{s2snn} is proposed to improve the performance of neural machine translation.
The same framework is then proposed to applying to the neural dialogue generation\cite{ncm}.
Based on the above Seq2Seq framework, the attention-based neural Seq2Seq model~\cite{att2014} is proposed to resolve the neural machine translation in the RNN-based encoder-decoder architecture.
The use of soft alignment and weight calculation over each time step improve the performance of the vanilla Seq2Seq model on machine translation and further adapt to dialogue generation.
Shang et al. (2015)~\cite{nrm} proposed a hybrid attentive mechanism to model the global and local relevance between input message and target response in short text conversation (single turn conversation).
Shao et al. (2017)~\cite{EMNLP17-1234} introduced an attentive model by adding a self-attention in decoding phase to improve the coherence and diversity of generated responses.
Yao et al. (2017)~\cite{EMNLP17-1232} proposed a two-stage approach which includes a cue words inference and a cue word guided response generation in an encoder-decoder framework.
Zhu et al. (2019)~\cite{zhu2019retrieval} utilized a retrieval model to enhance the generation of dialogue responses in an adversarial learning framework. 
Zhang et al. (2019)~\cite{zhang2019neural} proposed an initialization and adaptation framework to generate personalized dialogue responses. 
The LTS model~\cite{zhu2016learning} is utilized as a base model for the transfer learning process. 

To consider the use of dialogue history for diverse generation of conversational responses,
the HRED model~\cite{hred} is proposed to recurrently encode the dialogue history utterance by utterance and simultaneously use a RNN to encode the tokens inner utterance.
However, due to the gradient vanishing problem, they further proposed a vHRED~\cite{vhred} model to improve the relevance of generated response and the dialogue context by introducing a latent variable to model the posterior in training phase.
Due to the effect of attention mechanism on Seq2Seq learning~\cite{att2014}, the use of attention mechanism to modeling the contextual relevance~\cite{ms20171} is proposed to encode the inner and intra utterance relevance for multi turn dialogue generation.
Tian et al. (2017)~\cite{ACL17-2036} verified three measures of modeling the relevance between contextual utterances and response using sequential and hierarchical learning.
Recently, Zhang et al. (2019)~\cite{zhang2019recosa} proposed a self-attention based context encoder to model the relevance of contextual utterances and response for multi turn dialogue generation.
The encoder of the proposed ReCoSa model consists of a hierarchical attentive framework and a token level LSTM.
Besides the use of discriminate framework, recently, due to the advantages of generative adversarial net and variational autoencoder,
The SeqGAN model\cite{seqgan} is proposed to score a partially generated sequence by the reward function and MC search under the reinforcement learning framework for Seq2Seq learning tasks.
Zhao et al. (2017)~\cite{ACL17-1061} improves the variational autoencoder (VAE) by adding the response into the VAE in training phase to model the relevance of prior network and recognition network for diverse generation of dialogue.
Similar conclusions are also drawn by~\cite{ACL17-2080}.
Zhu et al. (2019)~\cite{zhu2019order} proposed a two-stage model to first predict a list of keywords, which are highly related to the generated response. 
And the second stage is to generate a response in the conditioning of the predicted keywords and the input message. 
The two-stage model is verified to better control the semantics and diversity of the generated responses than the end-to-end generation frameworks. 

As the success of reinforcement learning on modeling human-computer interactions, such as the AlphaGo~\cite{alphago} in the game of Go, some recent works have focused on modeling human-computer dialogue by not only considering the generation quality of current response, but also the future goal of the conversation~\cite{drl,dal,zhang2018exploring,ren2020crsal}.
To tackle the generation of generic and vague dialogue, a deep reinforcement learning approach~\cite{drl} is proposed with fine grained reward functions for multi turn dialogue generation.
An active learning approach~\cite{dal} is proposed to learn users' explicit feedback with a step of supervised learning for dialogue generation.
Zhang et al. (2018)~\cite{zhang2018exploring} proposed to model users' implicit feedback, such as opinion, sentiment and stalemate states, to reshape the reward function to generate multi turn coherent dialogue.
Zhu et al. (2020)~\cite{zhu2020counterfactual}  introduced a causal inference framework, SCM, to generate the counterfactual responses using an off-policy training approach.  
The counterfactual off-policy training is verified to be effective in improving the diversity and robustness of the dialogue model.  
In addition, in the view of a complete dialogue system, there are some other related works and challenges summarized by~\cite{huang2020challenges, fu2020context, chen2020aiis}. 

\section{The Proposed Approach}

\subsection{Preliminary} \label{pre}

\begin{figure*}[h]
\centering
\includegraphics[width=\linewidth]{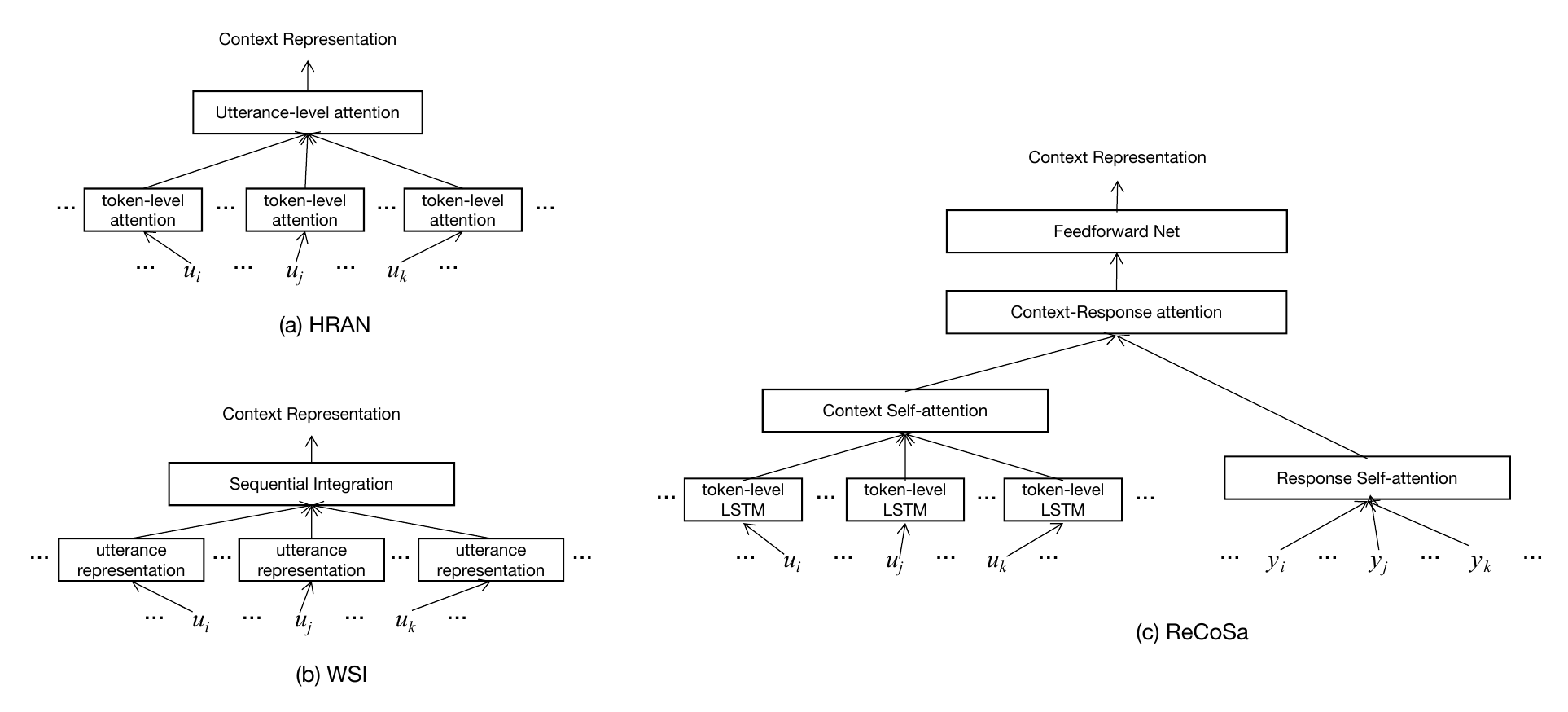}
\caption{The encoders of three prevalent frameworks, (a) HRAN~\cite{ms20171}, (b) WSI~\cite{ACL17-2036}  and (c) ReCoSa~\cite{zhang2019recosa} , for the generation of multi turn dialogue responses in open domain.}
\label{bases}
\end{figure*}

A general Seq2Seq learning-based neural network for multi turn conversation generation in open domain usually consists of an encoder and a decoder.
The encoder converts the dialogue utterances into a dense vector or a matrix to represent the semantics of the dialogue history in a multi turn dialogue session.
While, taking the output of the encoder as an input, the decoder then generates a response word by word.
In case of the generation of multi turn dialogue responses in this paper, the input message consists of a sequence of multi turn dialogue utterances.
Hence, the fundamental problem is how to represent the semantics of context in encoding phase.
Figure~\ref{bases} shows the three prevalent encoders in representing the semantics of multi turn dialogue context.

Here, $u_i$ denotes the $i$-th utterance in a multi turn dialogue.
$y_i$ represents the generated token of a response in time step $i$ by the decoder.
It is worth to note that the response can be either a gold response or a predicted response of last turn in training and test phases, respectively.
From Figure~\ref{bases}, we can see that all the three frameworks are taking the utterance-level representation as a basic unit of input for context modeling.
Meanwhile, the utterance-level representations are then hierarchically represented to obtain the context representation of a dialogue session through sequential integration~\cite{ACL17-2036} or attention~\cite{ms20171} (self-attention~\cite{zhang2019recosa} \footnote{Here, the content-response attention is utilized in the manner of self-attention. }).
Besides the common part of each framework, there are also some different parts among the three frameworks: %, which can be compared as follows:

\subsubsection*{$\bullet$~~Token-level Encoding: Bidirectional modeling vs. Unidirectional modeling}
Xing et al. (2017)~\cite{ms20171} employed a bidirectional GRU and a global attention mechanism to model word/token-level encodings.
While, Tian et al. (2017)~\cite{ACL17-2036} and Zhang et al. (2019)~\cite{zhang2019recosa} utilized a unidirectional GRU and LSTM to encode the word/token embedding, respectively.

\subsubsection*{$\bullet$~~Context Encoding: Attentive/Self-attentive modeling  vs. Sequential modeling}
Xing et al. (2017) \cite{ms20171} proposed to utilize a hierarchical attentive encoding framework to representing the context of multi turn dialogue.
Tian et al. (2017) \cite{ACL17-2036} proposed a weighted sequential integration (WSI) approach under the RNN framework to obtain the context representation.
Zhang et al. (2019)\cite{zhang2019recosa}  also introduced a hierarchical attentive mechanism for modeling the representation of context and response with self-attention respectively.  
An attention layer is utilized to model the interaction between context and response representations. 

\subsection{The Proposed Model} \label{model}

\begin{figure*}[h]
\centering
\includegraphics[width=\linewidth]{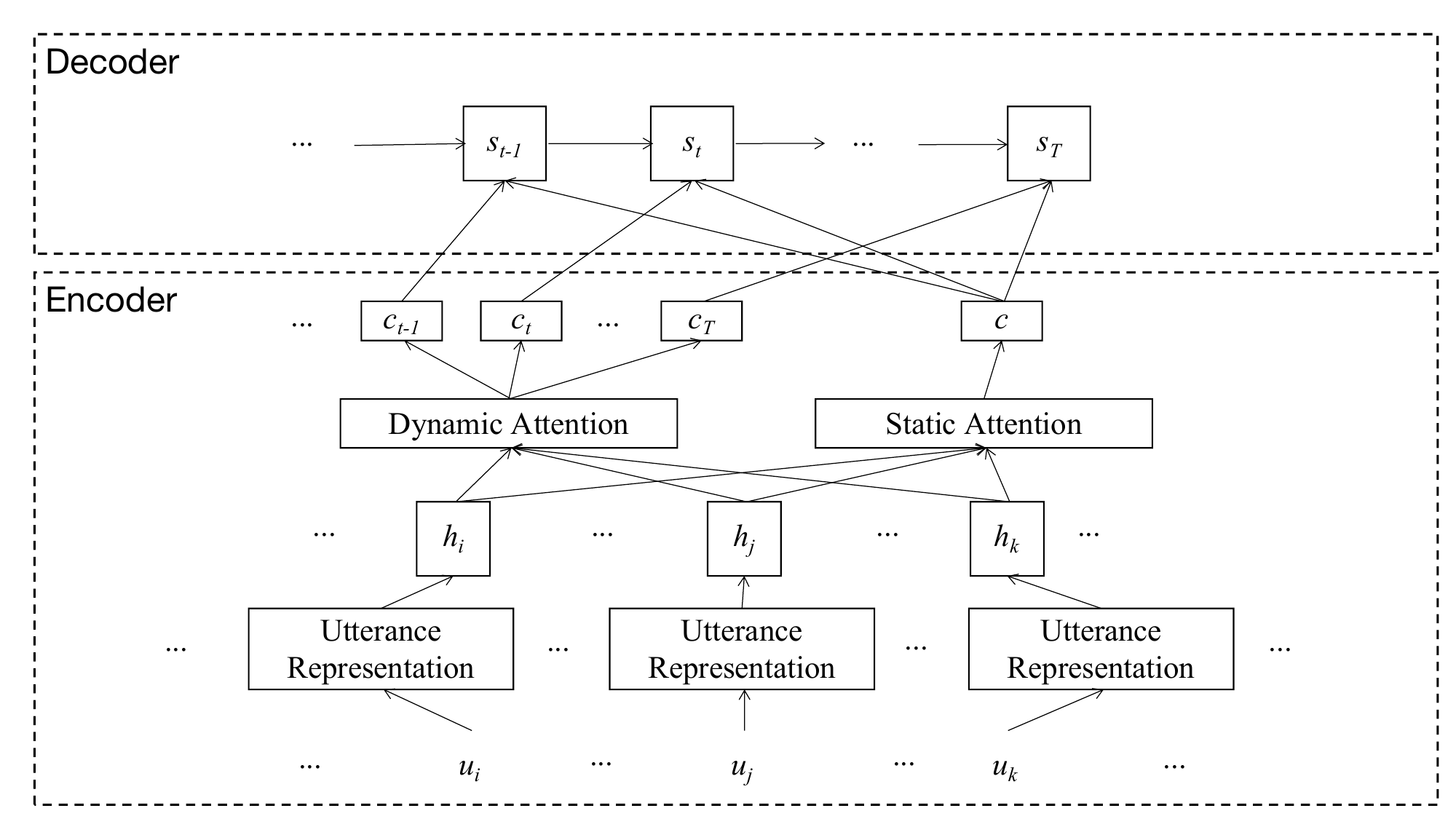}
\caption{The proposed attentive framework for the generation of open domain dialogues. Here, $u_*$ denotes the $*$-th utterance in a conversation. 
$h_*$ is the hidden state. $c_*$ and $c$ indicate the context representations that are obtained by the dynamic attention and static attention respectively. }
\label{framework}
\end{figure*}

The proposed attentive framework, which consists of a static and dynamic attention mechanism, for multi turn dialogue generation is based on the encoder-decoder architecture.
To obtain the context representation, the encoder of the proposed framework hierarchically encode the representation of the utterances in a dialogue session.
To obtain the utterance representation, we take the advantages of the three prevalent approaches to encoding the dialogue context information for the generation of multi turn dialogue responses~\cite{ms20171,ACL17-2036,zhang2019recosa}.
We present two mechanisms to obtain the representation of an utterance.
One is the GRU~\cite{chung2014empirical} model, which is recurrently modeling the embedding of words in each utterance.
The other is the Transformer~\cite{vaswani2017attention} encoding, which includes the multi-head self-attention and positional encoding.
To obtain the context representation in upper layer of the proposed model, the proposed attentive framework, which consists of a dynamic and a static attention, measures the weight of each utterance in representing the meaning of a dialogue session.
Figure~\ref{framework} shows the proposed framework.

In this paper, we model the inter-utterance representations to obtain the context representation of a multi turn dialogue in three measures, namely static, dynamic and hybrid attention, as shown in Figure~\ref{framework}.
Next, we will formally describe the three attentive mechanisms.

\subsubsection*{$\bullet$~~Static Attention}
As shown in Figure~\ref{framework}, the static attention mechanism calculates the importance of each utterance as $e_i$ or $\alpha_i$ ($i \in \{1,...,S\}$).
\begin{equation}\label{static-e}
e_i = V^T\text{tanh}(Wh_i+Uh_S)
\end{equation}
\begin{equation}\label{static-a}
\alpha_i = \frac{\text{exp}(e_i)}{\sum_i \text{exp}(e_i)}
\end{equation}
\begin{equation}\label{static-c}
c = \sum_i \alpha_i h_i
\end{equation}
Here, $h_i$ and $h_S$ denote the representations of the hidden state of the $i$-th and the last utterance in a dialogue session, respectively.
$V$, $W$ and $U$ are parameters.
We can see that once the weights of each utterance $\alpha_i$ ($i \in \{1,...,S\}$) are produced, they will be unchanged in the decoding process.
In decoding, the $t$-th hidden state $s_t$ can be calculated as follows:
\begin{equation}\label{static-decode}
s_t = f(y_{t-1}, s_{t-1}, c)
\end{equation}
Here, $y_{t-1}$ is the $t-1$-th output of the decoder and $s_{t-1}$ is the hidden state of $t$-$1$-th time step in decoding.
Notice that $y_0$ is set to be a special character and $s_0$ is initialized by $h_s$.
The generated response is thus represented as a sequence of $y_1,y_2,...,y_T$, where $T$ denotes the termination step.

\subsubsection*{$\bullet$~~Dynamic Attention}
Rather than the unchanged weights of each utterance in decoding phase, the dynamic attention maintains a weighting matrix and updates the weights of each utterance during the decoding process. 
The formal illustration of the dynamic attentive mechanism is as follows:
\begin{equation}\label{dynamic-e}
e_{i,t} = V^T\text{tanh}(Wh_i+Us_{t-1})
\end{equation}
\begin{equation}\label{dynamic-a}
\alpha_{i,t} = \frac{\text{exp}(e_{i,t})}{\sum_i \text{exp}(e_i)}
\end{equation}
\begin{equation}\label{dynamic-c}
c_t = \sum_i \alpha_{i,t} h_i
\end{equation}
Here, $V$, $W$ and $U$ are also parameters that are independent to those in the static attention.
$T$ denotes the transposition operation of $V$.
The $e_{i,t}$ and $\alpha_{i,t}$ are calculated in each time step $t$ of decoding.
The $t$-th hidden state $s_t$ in dynamic attention-based decoder can be calculated as follows:
\begin{equation}\label{dynamic-decode}
s_t = f(y_{t-1}, s_{t-1}, c_t)
\end{equation}

\subsubsection*{$\bullet$~~Hybrid Attention}
\label{secHyb}

The hybrid attention denotes the combination of the static and dynamic attentions.
There are two natural ways for the combination.
First is to concatenate the context vector $c$ in static attention-based encoding and the context vector $c_t$ in dynamic attention-based encoding.
Second is to sum up the context vector $c$ in static attention-based encoding and the context vector $c_t$ in dynamic attention-based encoding.
Equation (\ref{hybconcate}) and (\ref{hybsum}) denote the above two types of combination respectively.
\begin{equation}\label{hybconcate}
s_t = f(y_{t-1}, s_{t-1}, [c,c_t])
\end{equation}
\begin{equation}\label{hybsum}
s_t = f(y_{t-1}, s_{t-1}, c+c_t)
\end{equation}
Here, besides the $[c,c_t]$ and $c+c_t$, other notations are same to the above definition.

In addition,  we also try another two ways to explore the advantages of the two attention mechanisms. 
First is the linear interpolation of the context vector $c$ in static attention-based encoding and the context vector $c_t$ in dynamic attention-based encoding. 
Equation (\ref{hyblinear}) shows the state $s_t$ with linear interpolation. 
\begin{equation}\label{hyblinear}
s_t = f(y_{t-1}, s_{t-1}, \alpha \cdot c + \beta \cdot c_t)
\end{equation}
There are also three ways to fix the values of $\alpha$ and $\beta$. 
\begin{itemize}
\item We can empirically let both $\alpha$ and $\beta$ equals to 1, as shown in Equation (\ref{hybsum}).
\item We can let $\alpha$ and $\beta$ be the learnable variables in the multi turn dialogue generation process. 
\item We can also use attention mechanism to obtain the values of $\alpha$ and $\beta$. 
Here, we respectively calculate the cosine similarities of $c$, $c_t$ to the representation of hidden state in current time-step,  $s_{t-1}$. 
Thus $\alpha=\cos(c,s_{t-1})$ and $\beta=\cos(c_t,s_{t-1})$.
\end{itemize}

Second, we use element-wised pooling operation to combine the static and dynamic attentions. 
The ways of element-wised pooling of $c$ and $c_t$ include max-pooling and average pooling. 

\subsubsection*{$\bullet$~~Multi-head Attention}
According to the experimental results in~\cite{vaswani2017attention}, multi-head attentions can learn diverse features of the given input and achieve better performance.
Therefore, we extend the proposed ``single-head'' static and dynamic attentions to multi-head attentions. 
\begin{equation}\label{multihead}
  \begin{aligned}
    \operatorname{MultiHead}(Q, K, V) &=\text { Concat }\left(\operatorname{head}_{1}, \ldots, \text { head }_{\mathrm{h}}\right) W^{O} \\ 
    \text { where head }_{\mathrm{i}} &=\operatorname{g}\left(h, h_S, s_{t-1}\right)
  \end{aligned}
\end{equation}
As shown in Equation (\ref{multihead}), we use different parameters to calculate the values of static and dynamic attentions. 
Here, head$_{\mathrm{i}}=\text{g}(h, h_S, s_{t-1})$ denotes a set of static and dynamic attentions. 
$h$ and $h_S$ denote the hidden state of , respectively. 
$W^{O}$ is the parameter matrix for different static and dynamic attentions. 
The final representation is obtained by concatenating the multi-head attentions and then passing a linear layer. 

\subsubsection*{$\bullet$~~Token-level Information}
As shown in Figure~\ref{framework}, the proposed static and dynamic attentions only model the utterance-level information for multi turn dialogue generation. 
To verify the impact of token-level information, we further introduce the token representations in the proposed static and dynamic attention models. 
Concretely, first, the original input representations of utterances are substituted by the token embeddings so that the model can take the token-level attention for decoding. 
And the rest parts of the model are unchanged. 
Second, for each utterance, we concatenate the original utterance representation and the representation obtained by token-level attention. 
The model then takes the concatenated representation as the input of decoder. 

Noticed that the main difference between the proposed model and the above three state-of-the-art models are three folds: 
\begin{itemize}
\item First, we propose two utterance-level attentions for weighting the importance of each utterance in dialogue context, which is more simple in structure and has less number of parameters than the hierarchical attention approach.
\item Second, in the proposed approach, the weights of utterance in dialogue context are learned by two attention mechanisms from the data, which is more reasonable and flexible than the heuristic based approach.
\item Third, other utterance-level representation learning approaches, such as self-attention based utterance representation learning, can be integrated to the proposed attention-based dialogue encoding and response generation framework.
\end{itemize}

\section{Experimental Results}

\subsection{Experiment Settings}
\textbf{Datasets and Parameters:}
In this paper, two experimental datasets are chosen to verify the performance of multi turn dialogue generation.
One is the Ubuntu dataset~\cite{lowe-ubuntu}, which is extracted from the Ubuntu Internet Relayed Chat (IRC) channel and widely used for the generation of open domain dialogue responses~\cite{hred,vhred,AAAI1714571}.
We follow the data partition of train-test data in~\cite{AAAI1714571}.
As there is no development set in~\cite{AAAI1714571}, to construct a development set, in this paper, we randomly sample the same number of dialogue sessions to the test set from the training set.
The other is the Opensubtitles dataset which is introduced in~\cite{tiedemann2009} and used in~\cite{dpcm,drl}.
Here, in this paper, the Opensubtitles dataset used in the experiments is the same to that of~\cite{zhang2018context}. 
The statistics of the two datasets are shown in Table~\ref{data}.
The vocabulary consists of unique tokens in both two datasets. 
The hyper-parameters are listed in Table~\ref{parameters}.

\begin{table}[htbp]
\caption{The statistics of two experimental datasets. Avg is short for average. \# denotes the number. A sample in training, development and test set is a dialogue session.}
\label{data}
\begin{tabular}{l|rr}
\toprule
& \bf Ubuntu & \bf Opensubtitles \\
\midrule
Train size & 429,915 & 792,000 \\
Dev size & 18,920 & 8,000 \\
Test size & 18,920 & 8,000 \\
Vocabulary size & 155,490 & 91,405 \\
Avg \# of utterances per session & 7.5 & 10 \\
Avg \# of words per utterance & 12.3 & 7.5 \\
\bottomrule
\end{tabular}
\end{table}

\begin{table}[htbp]
\caption{Parameter setting.}
\label{parameters}
\begin{tabular}{l|cc}
\toprule
& \bf Static Attention & \bf Dynamic Attention \\
\midrule
Dimension of hidden state & 512 & 1,024 \\
\midrule
Padding length & \multicolumn{2}{c}{15}  \\
Dimension of word embedding &  \multicolumn{2}{c}{200}  \\
Initial learning rate and & \multicolumn{2}{c}{\multirow{2}{*}{0.001/10$^{-5}$}}  \\
weight\_decay &  \multicolumn{2}{c}{}  \\
Dropout parameter &  \multicolumn{2}{c}{0.5}  \\
Mini-batch size &  \multicolumn{2}{c}{80}  \\
Number of iterations &  \multicolumn{2}{c}{10}  \\
Number of Heads &  \multicolumn{2}{c}{4}  \\
\bottomrule
\end{tabular}
\end{table}

\textbf{Baselines:}
For experimental comparisons, we choose seven baseline approaches in multi turn dialogue generation task.
Five of them are state-of-the-art approaches, which are VHRED, CVAE, WSI, HRAN and ReCoSa.
The detailed descriptions of the baselines are as follows:
\begin{itemize}
  \item \textbf{LSTM:}
  For the sequence-to-sequence (Seq2Seq) based generation of multi turn dialogues, the simple and natural way is to directly use the LSTM to encode all the utterances in a session word by word and then generate a response in decoding phase.
  \item \textbf{HRED:}
  The first generation model for multi turn dialogue, which is proposed by~\cite{hred}. 
  The HRED model is a hierarchical recurrent encoder-decoder framework that can use RNN-based model to recurrently encode the token representations into an utterance representation.
  And then also recurrently encode each utterance representation to obtain the context (dialogue history) representation. 
  \item \textbf{VHRED:}
  An extended version of the HRED model.
  For the problem of generating generic responses in HRED model, the VHRED~\cite{vhred} model incorporates a stochastic latent variable at utterance level for encoding and decoding to improve the diversity of the generated dialogue responses.
  \item \textbf{CVAE:}
  The CVAE model is a conditional variational autoencoder (VAE) based approach, which is proposed by~\cite{ACL17-1061}.
  It learns to capture the semantics of the gold responses in training phase and generating diverse dialogue responses.
  \item \textbf{WSI}, \textbf{HRAN} and \textbf{ReCoSa} are proposed by~\cite{ACL17-2036}, \cite{ms20171} and~\cite{zhang2019recosa} respectively. 
  We will detail the descriptions and comparisons of them in Section~\ref{pre} and~\ref{model} and the architectures of the three models are shown in Figure~\ref{bases}.
\end{itemize}

\subsection{Evaluation and Results}
\subsubsection{Automatic Evaluation}
The automatic evaluation metrics for the generated dialogue responses in open domain is still an open problem.
The BLEU score~\cite{bleu} is a widely used in evaluating the performance of machine translation systems.
However, due to the difference between machine translation and dialogue generation, BLEU is not suitable for dialogue generation as the meanings of the referred responses are various so that they may share less common words.
Meanwhile, the perplexity is often used to evaluate the performance of language model, but is not performing well on evaluating the relevance between messages and responses in dialogue~\cite{nrm,drl}.

To address the above issues, in this paper, we use the embedding metric that is proposed by~\cite{hred} and also used in~\cite{vhred} for evaluating the relevance of the generated dialogue responses to the dialogue context.
Rather than measuring the similarity between a generated responses $\hat r$ and the ground-truth responses $r$ in token-level or $n$-gram level, the embedding metric measures their semantic similarity by matching the semantic representations.
There are three aspects in the embedding metric, namely $\bf Average$, $\bf Greedy$ and $\bf Extrema$.
\begin{itemize}
\item To calculate the $\bf Average$ score, it first runs the element-wise arithmetic average of word embeddings in $\hat r$ and $r$ and get two sentence-level representations $v_{\hat r_{avg}}$ and $v_{r_{avg}}$.
The $\bf Average$ score $S_{avg}$ is then calculated by the cosine similarity of $v_{\hat r_{avg}}$ and $v_{r_{avg}}$.
\item To obtain the $\bf Greedy$ score, it first calculates the cosine similarity between each word embedding in $\hat r$ and each word embeddings in $r$.
The same operation is also done from $r$ to $\hat r$. 
After that, each word $\hat w_i$ in $\hat r$ has a most similar word in $r$ and a corresponding cosine similarity score $\hat s_i$.
And each word $w_j$ in $r$ also has a most similar word in $\hat r$ and a corresponding cosine similarity score $s_j$.
The $\bf Greedy$ score $S_{grd}$ is then calculated as the following equation.
\begin{equation}
S_{grd}=\frac{1}{2}(\frac{1}{|\hat r|}\sum_{i=1}^{|\hat r|}{\hat s_i}+\frac{1}{|r|}\sum_{j=1}^{|r|} s_j)
\end{equation}
 \item To calculate the $\bf Extrema$ score, we can construct two embedding matrices $m_{\hat r}$ and $m_{r}$ by arranging all the word embeddings of $\hat r$ and $r$. 
 Here, the $i$-th column of $m_{\hat r}$ is the embedding of the $i$-th word in $\hat r$ and the same to that of $m_{r}$.
 For each row in $m_{\hat r}$ and $m_{r}$, we select the maximum value respectively and obtain two new sentence embeddings, $v_{\hat r_{ext}}$ and $v_{r_{ext}}$.
 The $\bf Extrema$ score equals to the cosine similarity of $v_{\hat r_{ext}}$ and $v_{r_{ext}}$. 
\end{itemize}

In addition, generating diverse responses is also an important aspect for a dialogue generation model. 
Therefore, we also evaluate the diversity of the generated dialogue responses using Distinct metric, which is proposed by~\cite{dpcm}. 
Here, we use the \textbf{Distinct-1} and \textbf{Distinct-2} to measure the diversity of generated responses in the levels of unigram and bigram, respectively. 

\subsubsection*{$\bullet$~~Main Results} Table~\ref{autoevaU} and \ref{autoevaO} show the results of automatic evaluations on Ubuntu and Opensubtitles datasets, respectively.
We can see that the proposed conversation generation model with static attention performs better in both $\bf Average$ and $\bf Greedy$ metrics than baselines on two dataset.
And the proposed conversation generation model with dynamic attention is good at generating diverse dialogue responses. 
Meanwhile, we can also see that the ReCoSa model outperforms other models in $\bf Extrema$ score in Ubuntu dataset.
And the HRAN model achieves the best performance in $\bf Greedy$ score in Opensubtitles dataset.  
The results demonstrate that our proposed model is good at learning the sentence-level and word (token)-level representations of context in the generation of dialogue responses. 
While the strongest baseline model, ReCoSa, tends to learn the prime content (extreme element) in an embedding representation. 

To compare the dynamic and static attentions, we find that the results in embedding metric and distinct metric are different with the proposed two attention-based dialogue generation models. 
For embedding metric, static attention performs better than the dynamic attention. 
However, the dynamic attention model outperforms the static attention model in distinct metric. 
The reason may be that the contextual representation of a dialogue history in dynamic attention model is variable in each time step in decoding phase.
It potentially increases the variety in each time-step of decoding process rather than given an unchanged contextual representation and does not ``disturb'' the decoding phase in static attention-based model.  
The dynamic attention-based model maintains a varying contextual representation so that it may lead to the generation of more diverse tokens than the static attention-based model in dialogue responses .
Meanwhile, the better performance on embedding metric reveals that the static attention-based model is good at fitting the semantics of golden dialogue responses. 

\begin{table}[htbp]
\caption{The results of automatic evaluation on Ubuntu dataset. Dynamic and Static are our proposed approaches and the other models are baselines. 
${\rightarrow}$ and ${\leftrightarrow}$ denote the use of unidirectional and bidirectional GRU in the proposed model to obtain utterance representations, respectively. 
$\dagger$ denotes the results pass the statistical significance test with $p<0.05$ over baselines.
The results in bold indicate the best performance.}
\label{autoevaU}
\begin{tabular}{l|ccccc}
\toprule
\multirow{2}{*}{\bf Models} & \multicolumn{5}{c}{\bf Ubuntu}  \\
& \bf Average & \bf Greedy & \bf Extrema & \bf Distinct-1 & \bf Distinct-2 \\
\midrule
LSTM    & 0.2300 & 0.1689 & 0.1574 & 0.31 & 1.54  \\
HRED    & 0.5770 & 0.4169 & 0.3914 & 0.61 & 3.09  \\
VHRED   & 0.5419 & 0.3839 & 0.3627 & 0.44 & 2.69  \\
CVAE    & 0.5672 & 0.3982 & 0.3689 & 0.49 & 2.77  \\
WSI     & 0.5775 & 0.4196 & 0.3893 & 0.51 & 3.00  \\
HRAN    & 0.5964 & 0.4139 & 0.3898 & 0.54 & 2.89  \\
ReCoSa  & 0.6027 & 0.4412 & \textbf{0.3916} & 0.62 & 3.19 \\ 
\midrule
Dynamic$_\leftrightarrow$ & 0.5884 & 0.4246 & 0.3782 & 0.53 & 3.09  \\
Dynamic$_\rightarrow$  & 0.6048 & 0.4507 & 0.3791 & \textbf{0.82}$^\dagger$ & \textbf{3.54}$^\dagger$  \\
Static$_\leftrightarrow$  & 0.6020 & 0.4407 & 0.3772 & 0.55 & 2.47  \\
Static$_\rightarrow$  & \textbf{0.6071}$^\dagger$ & \textbf{0.4535}$^\dagger$ & 0.3873 & 0.81 & 3.46  \\ 
\bottomrule
\end{tabular}
\end{table}

\begin{table}[htbp]
\caption{The results of automatic evaluation on Opensubtitles dataset. Dynamic and Static are our proposed approaches and the other models are baselines. 
${\rightarrow}$ and ${\leftrightarrow}$ denote the use of unidirectional and bidirectional GRU in the proposed model to obtain utterance representations, respectively. 
$\dagger$ denotes the results pass the statistical significance test with $p<0.05$ over baselines.
The results in bold indicate the best performance.}
\label{autoevaO}
\begin{tabular}{l|ccccc}
\toprule
\multirow{2}{*}{\bf Models} & \multicolumn{5}{c}{\bf Opensubtitles} \\
& \bf Average & \bf Greedy & \bf Extrema & \bf Distinct-1 & \bf Distinct-2  \\
\midrule
LSTM     & 0.5549 & 0.5029 & 0.3897 & 1.36 & 3.71 \\
HRED    & 0.5571 & 0.5033 & 0.3932 & 1.01 & 4.01 \\
VHRED   & 0.5248 & 0.4842 & 0.3556 & 1.57 & 4.42 \\
CVAE    & 0.4708 & 0.3390 & 0.3173 & 1.68 & 4.65 \\
WSI     & 0.5598 & 0.4964 & 0.3903 & 1.49 & 5.14 \\
HRAN    & 0.5617 & \textbf{0.5195} & 0.3898 & 1.08 & 5.10 \\
ReCoSa  & 0.6080 & 0.4681 & 0.4651 & 1.68 & 5.41 \\ 
\midrule
Dynamic$_\leftrightarrow$ & 0.6001 & 0.4531 & 0.4633 & 1.67 & 5.88 \\
Dynamic$_\rightarrow$  & 0.6125 & 0.4613 & 0.4634 & \textbf{1.69}$^\dagger$ & \textbf{6.01}$^\dagger$ \\
Static$_\leftrightarrow$  & 0.5989 & 0.4578 & 0.4781 & 1.36 & 4.99 \\
Static$_\rightarrow$  & \textbf{0.6138}$^\dagger$ & 0.4786 & \textbf{0.4864}$^\dagger$ & 1.54 & 5.08 \\ 
\bottomrule
\end{tabular}
\end{table}

To see the results of each model in both Table~\ref{autoevaU} and \ref{autoevaO}, we can see that the performance of the proposed static and dynamic attention-based models are more consistent than other baseline models. 
It also empirically reveals the robustness of the proposed static and dynamic attention-based models on different datasets. 
In addition, we find that the unidirectional GRU based models outperform the bidirectional GRU based models in the current experimental settings.

\begin{table}[htbp]
  \caption{The cases of generated responses by the proposed Static and Dynamic models with the unidirectional(${\rightarrow}$) and bidirectional(${\leftrightarrow}$) GRU on Ubuntu and Opensubtitles datasets.}\label{case-direction}
  \centering
      \begin{tabular}{l|l} 
      %\toprule
      \multicolumn{2}{l}{\textbf{Ubuntu}}  \\
      \toprule
      \multicolumn{2}{l}{Case 1} \\ 
      \midrule
      \multirow{3}*{Context}& is there any way to install tomcat 5.5 on 12.04 ? \\
                            & wasn't it tomcat55 package \\ 
                            & apologies , it doesn't seem to be available anymore \\ 
                            \midrule
   Golden                   & tried tomcat 5.5 gives me a regex output and 5 isn't there \\ 
   \midrule
   Dynamic$_\leftrightarrow$& I	think	it	's	just	the	one	that	comes	with	ubuntu	1.04  \\
   Dynamic$_\rightarrow$    & i	was	thinking	of	it	,	but	it	might	be	a	bug	in	launchpad  \\
   Static$_\leftrightarrow$ & i	have	a	problem	with	the	kernel	,	i	have	a	minimal	install	cd	,   \\
   Static$_\rightarrow$     & i	think	it	's	not	a	package	that	is	called	"	tomcat	"   \\ 
   \midrule
   \multicolumn{2}{l}{Case 2} \\ 
   \midrule
   \multirow{3}*{Context}& my dns is working good , I also cleared them . Delays still present \\
                            &  did you try disabling ipv6 with the boot option : ipv6 . \\
                            & trying , can we make the disable function working without restarting ? \\ 
                            \midrule
    Golden                        & may be able to use sysctl but not sure \\
    \midrule
    Dynamic$_\leftrightarrow$& yes	,	but	it	's	not	working	.	I	have	no	idea	what	you	mean   \\
    Dynamic$_\rightarrow$    & I don’t know , but I don’t think it ’s possible to do that .  \\
    Static$_\leftrightarrow$ & I	don't	know	,	sorry	.   \\
    Static$_\rightarrow$     & I	don't	know	.	You	might	try	to	ping	the	gateway	address	.   \\ 
    \midrule
    \multicolumn{2}{l}{\textbf{Opensubtitles}}  \\
    \toprule
    \multicolumn{2}{l}{Case 1} \\ 
    \midrule
    \multirow{3}*{Context}& What 's the matter with you ? \\
                            & You gotta be sitting ? \\
                            & Yeah , you 'll end up as a rug . \\ 
                            \midrule
   Golden                   & Come on , you 're holding up the works , you mug . \\ 
   \midrule
   Dynamic$_\leftrightarrow$& I	'm	not	a	good	boy	. \\
   Dynamic$_\rightarrow$    & I	'll	take	care	of	you	.  \\
   Static$_\leftrightarrow$ & I 'll be right back .   \\
   Static$_\rightarrow$     & You	're	a	fool	,	you	know	,	you	're	a	cinch	.   \\ 
   \midrule
   \multicolumn{2}{l}{Case 2} \\ \midrule
   \multirow{3}*{Context}& She 's a novel twist . \\
                            & Of course , I know you have no liking for Miss Hatton . \\ 
                            & She came in here , got nasty , I had to put her in her place . \\ 
                            \midrule
    Golden                   & Yes , I heard you had some slight misunderstanding \\ 
    \midrule
    Dynamic$_\leftrightarrow$ & I	'm	sorry	,	but	I	didn	't	have	to	tell	her	.  \\
    Dynamic$_\rightarrow$    & She	's	a	lot	.  \\
    Static$_\leftrightarrow$ & She	was	a	fraud	,	but	she	's	not	a	girl	.  \\
    Static$_\rightarrow$     & Yes , And	I	'm	not	interested	in	her	either	.  \\ 
    \bottomrule
    \end{tabular}
\end{table}

Table~\ref{case-direction} shows the examples of generated responses by the proposed static and dynamic attention-based models with the unidirectional and bidirectional GRU on the two experimental datasets. 
From Table~\ref{case-direction}, we can see that in general the diversity and relevance of the generated responses by unidirectional GRU models are better than that of the bidirectional GRU models, which is consistent to the results in Table~\ref{autoevaU} and \ref{autoevaO}. 
Meanwhile, the static attention with unidirectional GRU model has better performance than other models, especially on avoiding to generate generic responses. 
The reason may be that in the proposed architecture the utterance representations obtained by the GRU model in forward and backward directions are quite different, which may lead to an unstable representation of the encoded dialogue context. 

\subsubsection*{$\bullet$~~Hybrid Attention Results}
As described in Section~\ref{secHyb} hybrid attention, there are 5 ways to combine the proposed static and dynamic attentions to a hybrid attention. 
We use $concat$ and $sum$ to denote the concatenation and directly sum of the static and dynamic attentions, as shown in Equation (\ref{hybconcate}) and (\ref{hybsum}), respectively
$learnable$ and $attention$ indicate the two ways to fix the parameters of linear combination of the static and dynamic attentions, as shown in Equation (\ref{hyblinear}).
$max$ and $mean$ denote the element-wised max and average pooling of the static and dynamic attentions, respectively. 
As shown in Table~\ref{autoevaU} and \ref{autoevaO}, in tuning the experimental parameters,  we found that the embedding metric and the distinct metric have no obvious positive correlation. 
When the embedding metric achieve the best performance, the distinct metric can not be simultaneously optimal, and vice versa. 
Therefore, to show the results more comprehensively, in the hybrid attention experiments, we separately show the best performance of the embedding metric and distinct metric. 
The unidirectional GRU is used for obtaining utterance representation. 
In the rest of this paper, we will use the same way to show the experimental results. 

Table~\ref{hyb-autoU} and~\ref{hyb-autoO} show the experimental results of hybrid attention on Ubuntu and Opensubtitles datasets, respectively.
\begin{table}[htbp]
\caption{The results of automatic evaluation of the proposed hybrid model on Ubuntu dataset. $concat$, $sum$, $learnable$, $attention$, $max$ and $min$ denote the five ways of combining the static and dynamic attentions as shown in Section~\ref{secHyb}. Dist1/2 denotes the metrics of Distinct-1 and Distinct-2. The results in bold indicate the best performance.}
\label{hyb-autoU}
    \begin{tabular}{c|*4{c}|*4{c}} 
    \toprule
       \multirow{2}*{\textbf{Model}} & \multicolumn{4}{c|}{\textbf{Embedding Metrics Best}} & \multicolumn{4}{|c}{\textbf{Distinct Best}} \\
       ~ & \bf Average & \bf Greedy & \bf Extrema  & \bf Dist1/2 & \bf Average & \bf Greedy & \bf Extrema  & \bf Dist1/2 \\ 
       \midrule
      Dynamic   & 0.6048 & 0.4507 & 0.3791 & 0.82/3.54 & 0.5732 & 0.4398 & 0.3705 & 0.84/3.58 \\
      Static    & \textbf{0.6071} & \textbf{0.4535} & \textbf{0.3873} & 0.81/3.46 & 0.5724 & \textbf{0.4408} & \textbf{0.3750} & 0.91/3.99 \\ 
      \midrule
      $concat$    & 0.5934 & 0.4412 & 0.3770 & 0.82/3.59 & 0.5700 & 0.4348 & 0.3700 & 1.25/5.97 \\
      $sum$       & 0.5921 & 0.4425 & 0.3841 & 0.46/1.82 & 0.5658 & 0.4314 & 0.3670 & 1.24/5.68 \\
      $learnable$ & 0.5798 & 0.4473 & 0.3826 & \textbf{1.11}/4.63 & 0.5567 & 0.4270 & 0.3617 & 1.29/5.59 \\
      $attention$ & 0.5876 & 0.4442 & 0.3852 & 0.51/1.90 & 0.5627 & 0.4308 & 0.3664 & 1.28/\textbf{6.03} \\
      $max$       & 0.5849 & 0.4471 & 0.3859 & 1.08/\textbf{4.66} & 0.5623 & 0.4314 & 0.3685 & \textbf{1.32}/5.93 \\
      $mean$      & 0.5881 & 0.4445 & 0.3792 & 0.54/2.16 & \textbf{0.5743} & 0.4382 & 0.3701 & 1.25/5.75 \\ 
      \bottomrule
\end{tabular}
\end{table}

\begin{table}[htbp]
\caption{The results of automatic evaluation of the proposed hybrid model on Opensubtitles dataset. $concat$, $sum$, $learnable$, $attention$, $max$ and $min$ denote the five ways of combining the static and dynamic attentions as shown in Section~\ref{secHyb}. Dist1/2 denotes the metrics of Distinct-1 and Distinct-2. The results in bold indicate the best performance.}
\label{hyb-autoO}
    \begin{tabular}{c|*4{c}|*4{c}} 
    \toprule
       \multirow{2}*{\textbf{Model}} & \multicolumn{4}{c|}{\textbf{Embedding Metrics Best}} & \multicolumn{4}{|c}{\textbf{Distinct Best}} \\
       ~ & \bf Average & \bf Greedy & \bf Extrema  & \bf Dist1/2 & \bf Average & \bf Greedy & \bf Extrema  & \bf Dist1/2 \\ 
      \midrule
      Dynamic   & 0.6125 & 0.4613 & 0.4634 & 1.69/6.01 & 0.6018 & 0.4522 & 0.4539 & 2.40/9.01 \\
      Static    & \textbf{0.6138} & \textbf{0.4786} & \textbf{0.4864} & 1.54/5.08 & 0.5999 & 0.4470 & 0.4491 & 2.26/8.12 \\ 
      \midrule
      $concat$    & 0.6074 & 0.4624 & 0.4602 & 1.75/5.76 & 0.6056 & 0.4549 & 0.4590 & 2.51/9.41 \\
      $sum$       & 0.6093 & 0.4622 & 0.4627 & 1.86/6.16 & 0.6018 & 0.4513 & 0.4524 & 2.55/9.07 \\
      $learnable$ & 0.6108 & 0.4616 & 0.4594 & 1.69/5.34 & 0.6055 & 0.4544 & 0.4562 & 2.66/9.17 \\
      $attention$ & 0.6128 & 0.4634 & 0.4700 & 1.93/6.09 & 0.5969 & 0.4491 & 0.4491 & \textbf{2.67}/9.43 \\
      $max$       & 0.6092 & 0.4568 & 0.4611 & \textbf{2.49/8.72} & \textbf{0.6064} & \textbf{0.4574} & \textbf{0.4623} & 2.56/9.14 \\
      $mean$      & 0.6130 & 0.4633 & 0.4665 & 1.82/5.93 & 0.5998 & 0.4506 & 0.4534 & 2.65/\textbf{9.60} \\ 
      \bottomrule
\end{tabular}
\end{table}

We can see from Table~\ref{hyb-autoU} and~\ref{hyb-autoO} that in the setting of Embedding Metrics Best, the performance is decreased in all the combination ways on both Ubuntu and Opensubtitles datasets. 
However, the performance in distinct metric has gains by combining the static and dynamic attentions. 
It may be because that the introducing of dynamic attention can increase the diversity of the generated responses, but also make the semantics of generated responses deviate from the golden responses. 

\begin{table}[htbp]
  \caption{The cases of generated responses by the proposed hybrid model on Ubuntu dataset with different integration measures of static and dynamic attentions.}\label{case-ub-fusion}
  \centering
    \begin{tabular}{c|l} 
      \multicolumn{2}{l}{\textbf{Ubuntu}}  \\
      \toprule
      \multicolumn{2}{l}{Case 1} \\ \midrule
      \multirow{3}*{Context}& is there any way to install tomcat 5.5 on 12.04 ? \\
                            & wasn't it tomcat55 package \\ 
                            & apologies , it doesn't seem to be available anymore \\ 
                            \midrule
   Golden                   & tried tomcat 5.5 gives me a regex output and 5 isn't there \\ 
   \midrule
   Dynamic     & i	was	thinking	of	it	,	but	it	might	be	a	bug	in	launchpad  \\
   Static    & i	think	it	's	not	a	package	that	is	called	"	tomcat	" \\
   \midrule
   $concat$     & it	's	not	in	the	repos \\
   $sum$        & I	don't	think	it	's	a	good	idea	to	start	it	from	the	repositories	, \\
   $learnable$  & it	's	a	package	that	contains	the	package	name \\
   $attention$  & it	's	not	in	the	repos \\
   $max$        & I	think	it	's	just	a	bug	,	but	I	don't	know	how	to	fix \\
   $mean$       & it	's	not	in	the	repos \\ \midrule \midrule
  %  Line 272
      \multicolumn{2}{l}{Case 2} \\ \midrule
      \multirow{3}*{Context}& my dns is working good , I also cleared them . Delays still present \\
                            &  did you try disabling ipv6 with the boot option : ipv6 . \\
                            & trying , can we make the disable function working without restarting ? \\ 
                            \midrule
    Golden                        & may be able to use sysctl but not sure \\
    \midrule
    Dynamic    & I don’t know , but I don’t think it ’s possible to do that .   \\
    Static    & I	don't	know	.	You	might	try	to	ping	the	gateway	address	.   \\ 
    \midrule
    $concat$     & I	don't	know	.	I	don't	know	what	to	do	. \\
    $sum$        & I	don't	know	,	I	have	never	used	it	before \\
    $learnable$  & I	don't	know	,	I	haven't	used	it	myself	,	but	I	have	to	go \\
    $attention$  & I	think	it	's	not	working	. \\
    $max$        & I	don't	know	,	but	I	don't	think	it	's	possible	. \\
    $mean$       & I	don't	know	how	to	do	that	,	but	I	don't	know	how	to	set \\ 
    \bottomrule
    \end{tabular}
\end{table}

\begin{table}[htbp]
  \caption{The cases of generated responses by the proposed hybrid model on Opensubtitles dataset with different integration measures of static and dynamic attentions.}\label{case-op-fusion}
  \centering
    \begin{tabular}{c|l} 
      \multicolumn{2}{l}{\textbf{Opensubtitles}}  \\
      \toprule
      \multicolumn{2}{l}{Case 1} \\ \midrule
      \multirow{3}*{Context}& What 's the matter with you ? \\
                            & You gotta be sitting ? \\
                            & Yeah , you 'll end up as a rug . \\ 
                            \midrule
   Golden                   & Come on , you 're holding up the works , you mug . \\ 
   \midrule
   Dynamic    & I	'll	take	care	of	you	.  \\
   Static    & You	're	a	fool	,	you	know	,	you	're	a	cinch	.   \\ 
   \midrule
   $concat$     & I	'll	be	back	in	a	minute	. \\
   $sum$        & You	're	gonna	get	a	punch	. \\
   $learnable$  & Come	on	,	get	up	. \\
   $attention$  & You	're	a	peach	,	boy	. \\
   $max$        & I	'm	not	gonna	hurt	you	. \\
   $mean$       & You	're	not	going	to	be	a	big	shot	of	yourself	. \\ 
   \midrule 
  %  Line 418
      \multicolumn{2}{l}{Case 2} \\ 
      \midrule
      \multirow{3}*{Context}& She 's a novel twist . \\
                            & Of course , I know you have no liking for Miss Hatton . \\ 
                            & She came in here , got nasty , I had to put her in her place . \\ 
                            \midrule
    Golden                   & Yes , I heard you had some slight misunderstanding \\ 
    \midrule
    Dynamic    & She	's	a	lot	.  \\
    Static    & Yes , And	I	'm	not	interested	in	her	either	.  \\ 
    \midrule
    $concat$     & I	don	't	know	how	to	thank	you	. \\
    $sum$        & She	's	a	big	girl	,	she	's	a	gambler	. \\
    $learnable$  & I	don	't	know	. \\
    $attention$  & I	'm	sorry	,	but	I	don	't	think	so	. \\
    $max$        & She	's	a	local	girl	. \\
    $mean$       & I	didn	't	know	you	were	going	to	be	a	big	shot	. \\ 
    \bottomrule
    \end{tabular}
\end{table}

Table~\ref{case-ub-fusion} and~\ref{case-op-fusion} show the examples of generated responses by the proposed hybrid model with different integration measures of static and dynamic attentions on the Ubuntu and Opensubtitles datasets, respectively. 
From Table~\ref{case-ub-fusion} and~\ref{case-op-fusion}, we can see that the current ways of combing the static and dynamic attentions may not lead to an improvement in the relevance of generated responses. 
But from the actual case, we can see that the combination of the two attentions may result in generating more unseen tokens comparing to the responses generated by the single static and dynamic attentions. 

\subsubsection*{$\bullet$~~Multi-Head Attention Results}
To further see the impact of multi-head attention, we verify the performance with various number of heads on Ubuntu and Opensubtitles datasets, respectively, as shown in Table~\ref{multi-head-ub} and~\ref{multi-head-op}. 
\begin{table}[htbp]
  \caption{Experimental results on Ubuntu dataset using multi-head attention. \#heads denote the number of attention heads. Dist1/2 denotes the metrics of Distinct-1 and Distinct-2. The results in bold indicate the best performance.}\label{multi-head-ub}
  \centering
    \begin{tabular}{c|c|*4{c}|*4{c}} 
    \toprule
      \multirow{2}*{Model} & \multirow{2}*{\#heads} & \multicolumn{4}{c|}{Embedding Metrics Best} & \multicolumn{4}{|c}{Distinct Best} \\
      ~  & ~   & Average & Greedy & Extrema & Dist1/2 & Average & Greedy & Extrema & Dist1/2 \\ 
      \midrule
      \multirow{5}*{Dynamic} 
         &  1   & 0.5939 & 0.4477 & 0.3815 & 0.76/3.28 & 0.5580 & 0.4150 & 0.3633 & \textbf{1.57/8.08} \\
      ~  &  2   & 0.5844 & 0.4461 & 0.3878 & 0.54/2.17 & 0.5691 & 0.4379 & 0.3731 & 0.88/3.92 \\
      ~  &  4   & \textbf{0.6048} & \textbf{0.4507} & 0.3791 & \textbf{0.82/3.54} & 0.5732 & 0.4398 & 0.3705 & 0.84/3.58 \\
      ~  &  8   & 0.6032 & 0.4506 & 0.3885 & 0.51/2.14 & \textbf{0.5938} & 0.4395 & \textbf{0.3801} & 0.69/3.12 \\
      ~  &  12  & 0.5947 & 0.4483 & 0.3891 & 0.48/1.77 & 0.5705 & 0.4401 & 0.3716 & 0.62/2.28 \\
      ~  &  16  & 0.5939 & 0.4489 & \textbf{0.3914} & 0.41/1.55 & 0.5815 & \textbf{0.4422} & 0.3754 & 0.48/1.86 \\ 
      \midrule
      \multirow{5}*{Static} 
         &  1   & 0.5871 & 0.4443 & 0.3811 & 0.49/1.98 & 0.5599 & 0.4270 & 0.3638 & \textbf{1.30/5.82} \\
      ~  &  2   & 0.5880 & 0.4372 & 0.3830 & 0.59/2.52 & 0.5750 & 0.4362 & 0.3722 & 1.21/5.27 \\
      ~  &  4   & \textbf{0.6071} & \textbf{0.4535} & 0.3873 & \textbf{0.81/3.46} & 0.5724 & \textbf{0.4408} & 0.3750 & 0.91/3.99 \\
      ~  &  8   & 0.5900 & 0.4501 & 0.3891 & 0.71/2.80 & 0.5742 & 0.4378 & \textbf{0.3766} & 0.93/3.92 \\
      ~  &  12  & 0.5927 & 0.4503 & 0.3881 & 0.77/3.31 & 0.5717 & 0.4358 & 0.3701 & 0.87/3.87 \\
      ~  &  16  & 0.6038 & 0.4480 & \textbf{0.3899} & 0.31/1.18 & \textbf{0.5786} & 0.4396 & 0.3743 & 0.55/2.41 \\ 
      \bottomrule
    \end{tabular}
\end{table}

\begin{table}[htbp]
  \caption{Experimental results on Opensubtitles dataset using multi-head attention. \#heads denote the number of attention heads. Dist1/2 denotes the metrics of Distinct-1 and Distinct-2. The results in bold indicate the best performance.}\label{multi-head-op}
  \centering
    \begin{tabular}{c|c|*4{c}|*4{c}} 
    \toprule
      \multirow{2}*{Model} & \multirow{2}*{\#heads} & \multicolumn{4}{c|}{Embedding Metrics Best} & \multicolumn{4}{|c}{Distinct Best} \\
      ~  & ~   & Average & Greedy & Extrema & Dist1/2 & Average & Greedy & Extrema & Dist1/2 \\ 
      \midrule
      \multirow{5}*{Dynamic} 
         &  1   & 0.6062 & 0.4597 & 0.4637 & 2.18/7.54 & 0.6041 & 0.4535 & 0.4548 & \textbf{3.10/12.3} \\
      ~  &  2   & 0.6103 & 0.4553 & 0.4573 & \textbf{2.38/9.10} & \textbf{0.6052} & 0.4479 & 0.4540 & 2.33/9.34 \\
      ~  &  4   & \textbf{0.6125} & \textbf{0.4613} & 0.4634 & 1.69/6.01 & 0.6018 & 0.4522 & 0.4539 & 2.40/9.01 \\
      ~  &  8   & 0.6067 & 0.4547 & 0.4543 & 1.48/5.09 & 0.6000 & 0.4478 & 0.4534 & 1.53/5.38 \\
      ~  &  12  & 0.6003 & 0.4587 & \textbf{0.4638} & 0.42/1.37 & 0.5978 & \textbf{0.4673} & \textbf{0.4706} & 0.56/1.83 \\
      ~  &  16  & 0.5847 & 0.4583 & 0.4558 & 0.17/0.54 & 0.5847 & 0.4583 & 0.4558 & 0.17/0.54 \\ 
      \midrule
      \multirow{5}*{Static} 
         &  1   & 0.6025 & 0.4643 & 0.4690 & 1.84/5.85 & 0.6047 & 0.4580 & 0.4616 & \textbf{2.75/9.62} \\
      ~  &  2   & 0.6034 & 0.4570 & 0.4686 & \textbf{2.46/8.54} & 0.5971 & 0.4477 & 0.4508 & 2.46/8.77 \\
      ~  &  4   & \textbf{0.6138} & \textbf{0.4786} & \textbf{0.4864} & 1.54/5.08 & 0.5999 & 0.4470 & 0.4491 & 2.26/8.12 \\
      ~  &  8   & 0.6080 & 0.4604 & 0.4666 & 2.28/8.30 & \textbf{0.6080} & \textbf{0.4604} & \textbf{0.4666} & 2.28/8.30 \\
      ~  &  12  & 0.6026 & 0.4627 & 0.4625 & 1.32/4.42 & 0.5938 & 0.4492 & 0.4450 & 2.11/7.71 \\
      ~  &  16  & 0.6120 & 0.4587 & 0.4717 & 0.89/2.68 & 0.6036 & 0.4582 & 0.4581 & 1.40/4.86 \\ 
      \bottomrule
    \end{tabular}
\end{table}

We can draw the following conclusion from Table~\ref{multi-head-ub} and~\ref{multi-head-op}. 
\begin{itemize}
\item In sum, the performance of multi-head attention is better than single-head attention. 
However, the performance is not linearly increasing with the number of attention heads. 
Increasing the number of heads leads to increase the learning difficulty of model so that to decrease the performance.  
\item As the increasing number of attention heads, the performance on distinct will decrease. 
This may be caused by the way of combination of multi-head attentions. 
In order to obtain a more stable representation of intermediate layer, each attention head is concatenated and then passing a linear layer, which may lead to decrease the diversity of generated responses. 
In addition, recent research work~\cite{sun2020generating} also reveals that directly sampling the representation of a head is better than the combining the representations of multi-head in generation diversity. 
However, it also leads to a worse generation quality of responses. 
The multi-head attention results also indicate the inconsistency of the embedding metric and the distinct metric in evaluating dialogue response generation task. 
\end{itemize}

\begin{table}[htbp]
  \caption{The cases of generated responses on Ubuntu dataset with different number of attention heads..}\label{case-ub-head}
  \centering
      \begin{tabular}{l|l} 
      \multicolumn{2}{l}{\textbf{Ubuntu}}  \\
      \toprule
      \multicolumn{2}{l}{Case 1} \\ \midrule
      \multirow{3}*{Context}& is there any way to install tomcat 5.5 on 12.04 ? \\
                            & wasn't it tomcat55 package \\ 
                            & apologies , it doesn't seem to be available anymore \\ \midrule
   Golden                   & tried tomcat 5.5 gives me a regex output and 5 isn't there \\ \midrule
   Dynamic 1-head     & i	think	it	's	a	bug , and you can	have	a	look	at	the	wiki \\
   Dynamic 4-head     & i was thinking of it , but it might be a bug in launchpad \\
   Dynamic 8-head     & I	don't	know	what	to	do	with	it \\
   Dynamic 16-head    & i	think	it	's	not	working \\ \midrule
   Static 1-head      & no	,	it	's	not	the	same \\
   Static 4-head      & i think it ’s not a package that is called " tomcat "  \\
   Static 8-head      & no	,	it	's	not	a	package  \\
   Static 16-head     & np  \\ \midrule \midrule
  %  Line 272
      \multicolumn{2}{l}{Case 2} \\ \midrule
      \multirow{3}*{Context}& my dns is working good , I also cleared them . Delays still present \\
                            &  did you try disabling ipv6 with the boot option : ipv6 . \\
                            & trying , can we make the disable function working without restarting ? \\ 
                            \midrule
    Golden                        & may be able to use sysctl but not sure \\
                            \midrule
    Dynamic 1-head     & I don’t think so. I have to go to bed. \\
    Dynamic 4-head     & I	don't	know	,	but	I	don't	think	it	's	possible	to	do	that	. \\
    Dynamic 8-head     & I	have	no	idea	,	I	don't	have	any	other	devices	that	I	use	. \\
    Dynamic 16-head    & I	don't	know	what	you	mean	by	default \\ \midrule
    Static 1-head      & yes	,	it	's	a	static	address	,	but	it	's	not	a	dns	issue \\
    Static 4-head      & I don’t know . You might try to ping the gateway address .  \\
    Static 8-head      & no	,	it	's	not	a	router	.  \\
    Static 16-head     & I	don't	know	what	I	do	.  \\ \bottomrule
    \end{tabular}
\end{table}

\begin{table}[htbp]
  \caption{The cases of generated responses on Opensubtitles dataset with different number of attention heads.}\label{case-op-head}
  \centering
    \begin{tabular}{l|l} 
    \multicolumn{2}{l}{\textbf{Opensubtitles}}  \\
    \toprule
      \multicolumn{2}{l}{Case 1} \\ \midrule
      \multirow{3}*{Context}& What 's the matter with you ? \\
                            & You gotta be sitting ? \\
                            & Yeah , you 'll end up as a rug . \\ \midrule
   Golden                   & Come on , you 're holding up the works , you mug . \\ \midrule
   Dynamic 1-head     & Come	on	. \\
   Dynamic 4-head     & I’ll take care of you. \\
   Dynamic 8-head     & You	got	a	little	bit	of	getting	in	the	water	. \\
   Dynamic 16-head    & You	. \\ \midrule
   Static 1-head      & I	'll	take	care	of	you	. \\
   Static 4-head      & You’re a fool , you know , you’re a cinch.  \\
   Static 8-head      & I	'm	gonna	get	you	a	little	more	time	.  \\
   Static 16-head     & Come	on	,	boys	.  \\ \midrule \midrule
  %  Line 418
      \multicolumn{2}{l}{Case 2} \\ \midrule
      \multirow{3}*{Context}& She 's a novel twist . \\
                            & Of course , I know you have no liking for Miss Hatton . \\ 
                            & She came in here , got nasty , I had to put her in her place . \\ \midrule
    Golden                   & Yes , I heard you had some slight misunderstanding \\ \midrule
    Dynamic 1-head     & I	was	just	thinking	about	going	abroad	. \\
    Dynamic 4-head     & She’s a lot. \\
    Dynamic 8-head     & And	I	have	to	tell	her	that	I	should	have	her	to	her	. \\
    Dynamic 16-head    & She	's	a	lot	.\\ \midrule
    Static 1-head      & I	don	't	know	. \\
    Static 4-head      & Yes , And I ’m not interested in her either .  \\
    Static 8-head      & I	was	just	thinking	of	her	.  \\
    Static 16-head     & She	's	a	nice	girl	.  \\ \bottomrule
    \end{tabular}
\end{table}

Table~\ref{case-ub-head} and~\ref{case-op-head} show the examples of generated responses by the proposed Static and Dynamic models with different number of attention heads on the Ubuntu and Opensubtitles datasets, respectively. 
From Table~\ref{case-ub-head} and~\ref{case-op-head} , we can see that in general the responses generated by 4-head attentions have better relevance than other number of attentions. 
The observation of cases is consistent to the experimental results on embedding metric in Table~\ref{multi-head-ub} and~\ref{multi-head-op}. 
The reason may be that less number of attention heads may lead to insufficient representation of features. 
However, more number of attention heads may also cause the overfitting of models.

\subsubsection*{$\bullet$~~Token-level Information Results}
Previous experiments of the proposed static and dynamic attention models are based on the utterance-level representations. 
In this section, we will further explore the performance of adding token-level information as an enhancement for multi turn dialogue generation. 
In experiments, we first replace each utterance representation to its corresponding token representations and leave the rest parts of the static and dynamic models unchanged. 
Second, the representations of the dialogue context/history, which are obtained by the original utterance-level model and the token-level model, are concatenated for decoding.
Table~\ref{token-levelU} and~\ref{token-levelO} show the experimental results with token-level information on Ubuntu and Opensubtitles datasets, respectively.  

\begin{table}[htbp]
  \caption{The results of automatic evaluation of proposed models with token-level information on Ubuntu dataset. TA indicates the token-level attention. Dist1/2 denotes the metrics of Distinct-1 and Distinct-2. The results in bold indicate the best performance.}\label{token-levelU}
  \centering
    \begin{tabular}{c|*4{c}|*4{c}} 
    \toprule
      \multirow{2}*{Model} & \multicolumn{4}{c|}{Embedding Metrics Best} & \multicolumn{4}{|c}{Distinct Best} \\
       ~ & Average & Greedy & Extrema & Dist1/2 & Average & Greedy & Extrema & Dist1/2 \\ 
      \midrule
      Dynamic   & 0.6048 & 0.4507 & 0.3791 & 0.82/3.54 & 0.5732 & 0.4398 & 0.3705 & 0.84/3.58 \\
      Static    & \textbf{0.6071} & \textbf{0.4535} & \textbf{0.3873} & 0.81/3.46 & 0.5724 & \textbf{0.4408} & 0.3750 & 0.91/3.99 \\
      Dynamic+TA  & 0.5928 & 0.4490 & 0.3789 & 0.70/2.94 & \textbf{0.5797} & 0.4394 & \textbf{0.3802} & \textbf{1.54/7.59}  \\
      Static+TA   & 0.5807 & 0.4437 & 0.3807 & \textbf{1.15/5.12} & 0.5687 & 0.4377 & 0.3691 & 1.27/5.91 \\ 
      \bottomrule
    \end{tabular}
 \end{table}

\begin{table}[htbp]
  \caption{The results of automatic evaluation of proposed models with token-level information on Opensubtitles dataset. TA indicates the token-level attention. Dist1/2 denotes the metrics of Distinct-1 and Distinct-2. The results in bold indicate the best performance.}\label{token-levelO}
  \centering
    \begin{tabular}{c|*4{c}|*4{c}} 
    \toprule
      \multirow{2}*{Model} & \multicolumn{4}{c|}{Embedding Metrics Best} & \multicolumn{4}{|c}{Distinct Best} \\
       ~ & Average & Greedy & Extrema & Dist1/2 & Average & Greedy & Extrema & Dist1/2 \\ 
      \midrule
      Dynamic   & 0.6125 & 0.4613 & 0.4634 & 1.69/6.01 & 0.6018 & 0.4522 & 0.4539 & 2.40/9.01 \\
      Static    & \textbf{0.6138} & \textbf{0.4786} & \textbf{0.4864} & 1.54/5.08 & 0.5999 & 0.4470 & 0.4491 & 2.26/8.12 \\
      Dynamic+TA& 0.6082 & 0.4537 & 0.4605 & \textbf{2.65/10.2} & \textbf{0.6044} & \textbf{0.4528} & \textbf{0.4582} & \textbf{3.05/11.71} \\
      Static+TA & 0.6093 & 0.4590 & 0.4656 & 1.83/5.76 & 0.6040 & 0.4517 & 0.4511 & 2.73/9.65 \\ 
      \bottomrule
    \end{tabular}
 \end{table}

From Table~\ref{token-levelU} and~\ref{token-levelO}, we can find that:
\begin{itemize}
\item Adding the token-level information will decrease the performance of embedding metric. 
The reasons may be that on the one hand, the utterance-level information includes enough semantics.
On the other hand, in order to avoid the instability in training process, the hierarchical encoder will filter out the redundant information, which is introduced by the token-level attention scheme. 
\item The performance of distinct metric is increasing after adding the token-level information. 
The reasons may be that first, the introducing of token-level attention as an external information increases the randomness of decoding process. 
Second, to compare the static and dynamic attentions, we find that the improvements of the dynamic attention model is larger than that of the static attention model. 
To sum, we can find two points that impact the diversity of the generation of dialogue responses. 
One is dynamically choosing the input information in decoding phase. 
The other is the introducing of external information, such as token-level information in the proposed model. 
\end{itemize}

\begin{table}[htbp]
  \caption{The cases of generated responses by the proposed Static and Dynamic models with token-level information on Ubuntu and Opensubtitles datasets.}\label{case-TA}
  \centering
    \begin{tabular}{l|l} 
      \multicolumn{2}{l}{\textbf{Ubuntu}}  \\
      \toprule
      \multicolumn{2}{l}{Case 1} \\ 
      \midrule
      \multirow{3}*{Context}& is there any way to install tomcat 5.5 on 12.04 ? \\
                            & wasn't it tomcat55 package \\ 
                            & apologies , it doesn't seem to be available anymore \\ 
                            \midrule
    Golden                   & tried tomcat 5.5 gives me a regex output and 5 isn't there \\ 
    \midrule
   Dynamic     & i	was	thinking	of	it	,	but	it	might	be	a	bug	in	launchpad  \\
   Dynamic + TA             & it	's	not	in	the	repos	,	it	's	just	not	working   \\
   Static    & i	think	it	's	not	a	package	that	is	called	"	tomcat	" \\    
   Static + TA              & i	have	no	idea	,	i	just	installed	it	from	the	ppa   \\ 
    \midrule
   \multicolumn{2}{l}{Case 2} \\ 
   \midrule
   \multirow{3}*{Context}& my dns is working good , I also cleared them . Delays still present \\
                            &  did you try disabling ipv6 with the boot option : ipv6 . \\
                            & trying , can we make the disable function working without restarting ? \\ 
                            \midrule
    Golden                        & may be able to use sysctl but not sure \\
    \midrule
    Dynamic    & I don’t know , but I don’t think it ’s possible to do that . \\
    Dynamic + TA             & I	don't	know	,	I	don't	use	it	myself   \\
    Static    & I	don't	know	.	You	might	try	to	ping	the	gateway	address	.   \\ 
    Static + TA              & I	don't	know	,	I	haven't	tried	it	.   \\ 
    \midrule    
    \multicolumn{2}{l}{\textbf{Opensubtitles}}  \\
    \midrule
    \multicolumn{2}{l}{Case 1} \\ 
    \midrule
    \multirow{3}*{Context}& What 's the matter with you ? \\
                            & You gotta be sitting ? \\
                            & Yeah , you 'll end up as a rug . \\ 
                            \midrule
   Golden                   & Come on , you 're holding up the works , you mug . \\ 
   \midrule
   Dynamic    & I	'll	take	care	of	you	.  \\
   Dynamic + TA             & I	'm	not	kidding	. \\
   Static    & You	're	a	fool	,	you	know	,	you	're	a	cinch	.   \\ 
   Static + TA              & You	're	gonna	be	a	bunch	of	assholes	. \\ 
   \midrule
   \multicolumn{2}{l}{Case 2} \\ 
   \midrule
   \multirow{3}*{Context}& She 's a novel twist . \\
                            & Of course , I know you have no liking for Miss Hatton . \\ 
                            & She came in here , got nasty , I had to put her in her place . \\ 
                            \midrule
    Golden                   & Yes , I heard you had some slight misunderstanding \\ 
    \midrule
    Dynamic    & She	's	a	lot	.  \\
    Dynamic + TA             & I	've	got	to	get	rid	of	her	.  \\
    Static    & Yes , And	I	'm	not	interested	in	her	either	.  \\ 
    Static + TA              & She	was	a	fraud	,	and	she	gave	me	a	box	of	sugar	. \\ 
   \bottomrule
    \end{tabular}
\end{table}

Table~\ref{case-TA} shows the examples of generated responses by the proposed Static and Dynamic models with token-level information on the Ubuntu and Opensubtitles datasets, respectively. 
From Table~\ref{case-TA} , we can see that on one hand the integration of token-level information can improve the relevance of the generated response comparing to the vanilla Static and Dynamic models. 
The reason may be that the integrating of token-level information helps to maintain the semantics of the input dialogue context so that to obtain better representations for decoding phase.  
On the other hand, we can see from the cases that the integration of token-level information leads to generate more diverse tokens in responses, which is consistent to the improvements of distinct metric shown in Table~\ref{token-levelU} and~\ref{token-levelO}.  
%======================================================

\subsubsection*{$\bullet$~~Pretraining Enhanced Results}
In this section, we want to verify the impact of pre-trained language models on the proposed contextual dialogue generation framework.

We tried to integrate the GPT2~\footnote{\url{https://s3.amazonaws.com/models.huggingface.co/bert/gpt2-pytorch_model.bin}} model into the proposed static and dynamic attention-based models. 
Thanks to the large-scale parameters and the Transformer-based model architecture, one of the strongest capabilities of the pretrained models is precisely representing the semantics of words. 
Therefore, in considering not to make fundamental changes to the proposed framework, we use a simple way to take the advantages of GPT2 to the proposed attention-based dialogue generation framework. 
In detail, we first use the tokenizer of the GPT2 to transfer the tokens in an input (a sequence of utterances, e.g., dialogue context) to the embeddings of the GPT2 model.
Second, these token embeddings are encoded by the proposed static and dynamic attention-based models. 
At last, in decoding phase, the generated list of token IDs is then realized by the GPT2 model into a dialogue response.
It is worth to note that in the first step, if the input tokens are not able to match any tokens in the GPT2 model, these tokens will be recognized as UNK tokens. 

Table~\ref{pretrainU} and~\ref{pretrainO} show the experimental results of integrating the GPT2 into the proposed Dynamic and Static models on the Ubuntu and Opensubtitles datasets, respectively. 

\begin{table}[htbp]
\caption{The results of automatic evaluation of the proposed hybrid model with GPT2 on Ubuntu dataset. $concat$, $sum$, $learnable$, $attention$, $max$ and $min$ denote the five ways of combining the static and dynamic attentions as shown in Section~\ref{secHyb}. Dist1/2 denotes the metrics of Distinct-1 and Distinct-2. The $\uparrow$ indicates the improvement of performance by integrating the GPT2 representation.  The results in bold indicate the best performance.}
\label{pretrainU}
    \begin{tabular}{c|*4{c}|*4{c}} 
    \toprule
       \multirow{2}*{Model} & \multicolumn{4}{c|}{Embedding Metrics Best} & \multicolumn{4}{|c}{Distinct Best} \\
       ~ & Average & Greedy & Extrema & Dist1/2 & Average & Greedy & Extrema & Dist1/2 \\ 
       \midrule 
       \midrule
      Dynamic   & 0.6048 & 0.4507 & 0.3791 & 0.82/3.54 & 0.5732 & 0.4398 & 0.3705 & 0.84/3.58 \\
      +GPT2   & 0.6094 $\uparrow$ & 0.4461 & \textbf{0.3980} $\uparrow$ & 0.48/1.68 & 0.5936 $\uparrow$ & \textbf{0.4503} $\uparrow$ & \textbf{0.3851} $\uparrow$ & 1.25 $\uparrow$/4.86 $\uparrow$ \\ 
      \midrule
      Static    & 0.6071 & \textbf{0.4535} & 0.3873 & 0.81/3.46 & 0.5724 & 0.4408 & 0.3750 & 0.91/3.99 \\
      +GPT2    & 0.6098 $\uparrow$ & 0.4484 & 0.3857 & 1.05 $\uparrow$/3.88 $\uparrow$ & 0.5862 $\uparrow$ & 0.4405 & 0.3783 $\uparrow$ & \textbf{1.87} $\uparrow$/\textbf{8.18} $\uparrow$ \\ 
      \midrule 
      \midrule
      $concat$    & 0.5934 & 0.4412 & 0.3770 & 0.82/3.59 & 0.5700 & 0.4348 & 0.3700 & 1.25/5.97 \\
      +GPT2     & 0.6109 $\uparrow$ & 0.4404 & 0.3976 $\uparrow$ & 0.54/1.84 & 0.5882 $\uparrow$ & 0.4463 $\uparrow$ & 0.3827 $\uparrow$ & 1.32 $\uparrow$/5.18 \\ 
      \midrule
      $sum$       & 0.5921 & 0.4425 & 0.3841 & 0.46/1.82 & 0.5658 & 0.4314 & 0.3670 & 1.24/5.68 \\
      +GPT2     & 0.6063 $\uparrow$ & 0.4433 $\uparrow$ & 0.3889 $\uparrow$ & 0.68 $\uparrow$/2.44 $\uparrow$ & 0.5830 $\uparrow$ & 0.4425 $\uparrow$ & 0.3806 $\uparrow$ & 1.36 $\uparrow$/5.19 \\ 
      \midrule
      $learnable$ & 0.5798 & 0.4473 & 0.3826 & \textbf{1.11}/4.63 & 0.5567 & 0.4270 & 0.3617 & 1.29/5.59 \\
      +GPT2     & 0.5986 $\uparrow$ & 0.4304 & 0.3847 $\uparrow$ & 0.99/3.48 & 0.5839 $\uparrow$ & 0.4448 $\uparrow$ & 0.3774 $\uparrow$ & 1.26/4.76 \\ 
      \midrule
      $attention$ & 0.5876 & 0.4442 & 0.3852 & 0.51/1.90 & 0.5627 & 0.4308 & 0.3664 & 1.28/6.03 \\
      +GPT2     & 0.6119 $\uparrow$ & 0.4529 $\uparrow$ & 0.3922 $\uparrow$ & 0.81 $\uparrow$/2.83 $\uparrow$ & 0.5868 $\uparrow$ & 0.4345 $\uparrow$ & 0.3813 $\uparrow$ & 0.99/3.48 \\ 
      \midrule
      $max$       & 0.5849 & 0.4471 & 0.3859 & 1.08/\textbf{4.66} & 0.5623 & 0.4314 & 0.3685 & 1.32/5.93 \\
      +GPT2     & \textbf{0.6144} $\uparrow$ & 0.4512 $\uparrow$ & 0.3921 $\uparrow$ & 0.75/2.67 & 0.5864 $\uparrow$ & 0.4437 $\uparrow$ & 0.3814 $\uparrow$ & 1.25/4.76 \\ 
      \midrule
      $mean$      & 0.5881 & 0.4445 & 0.3792 & 0.54/2.16 & 0.5743 & 0.4382 & 0.3701 & 1.25/5.75 \\
      +GPT2     & 0.6134 $\uparrow$ & 0.4495 $\uparrow$ & 0.3970 $\uparrow$ & 0.51/1.78 & \textbf{0.5958} $\uparrow$ & 0.4478 $\uparrow$ & 0.3825 $\uparrow$ & 1.24/4.92\\ 
      \bottomrule
\end{tabular}
\end{table}

\begin{table}[htbp]
\caption{The results of automatic evaluation of the proposed hybrid model with GPT2 on Opensubtitles dataset. $concat$, $sum$, $learnable$, $attention$, $max$ and $min$ denote the five ways of combining the static and dynamic attentions as shown in Section~\ref{secHyb}. Dist1/2 denotes the metrics of Distinct-1 and Distinct-2. The $\uparrow$ indicates the improvement of performance by integrating the GPT2 representation.  The results in bold indicate the best performance.}
\label{pretrainO}
    \begin{tabular}{c|*4{c}|*4{c}} 
    \toprule
       \multirow{2}*{Model} & \multicolumn{4}{c|}{Embedding Metrics Best} & \multicolumn{4}{|c}{Distinct Best} \\
       ~ & Average & Greedy & Extrema & Dist1/2 & Average & Greedy & Extrema & Dist1/2 \\ 
       \midrule 
       \midrule
       Dynamic   & 0.6125 & 0.4613 & 0.4634 & 1.69/6.01 & 0.6018 & 0.4522 & 0.4539 & 2.40/9.01 \\
       +GPT2     & \textbf{0.6189} $\uparrow$ & 0.4601 & 0.4662 $\uparrow$ & 1.07/3.22 & 0.6001 & 0.4578 $\uparrow$ & 0.4589 $\uparrow$ & \textbf{4.20} $\uparrow$/10.52 $\uparrow$ \\ 
       \midrule
       Static    & 0.6138 & \textbf{0.4786} & 0.4864 & 1.54/5.08 & 0.5999 & 0.4470 & 0.4491 & 2.26/8.12 \\
       +GPT2     & 0.6163 $\uparrow$ & 0.4572 & \textbf{0.4899} $\uparrow$ & 0.80/2.36 & 0.5997 & 0.4506 $\uparrow$ & 0.4515 $\uparrow$ & 3.13 $\uparrow$/\textbf{10.91} $\uparrow$ \\ 
       \midrule 
       \midrule       
       $concat$    & 0.6074 & 0.4624 & 0.4602 & 1.75/5.76 & 0.6056 & 0.4549 & 0.4590 & 2.51/9.41 \\
       +GPT2     & 0.6150 $\uparrow$ & 0.4512 & 0.4644 $\uparrow$ & 0.63/1.84 & 0.6030 & 0.4523 & 0.4580 & 2.79 $\uparrow$/10.23 $\uparrow$ \\ 
       \midrule
       $sum$       & 0.6093 & 0.4622 & 0.4627 & 1.86/6.16 & 0.6018 & 0.4513 & 0.4524 & 2.55/9.07 \\
       +GPT2     & 0.6104 $\uparrow$ & 0.4533 & 0.4659 $\uparrow$ & 0.47/1.40 & 0.6025 $\uparrow$ & 0.4495 & 0.4615 $\uparrow$ & 3.32 $\uparrow$/10.12 $\uparrow$ \\ 
       \midrule
       $learnable$ & 0.6108 & 0.4616 & 0.4594 & 1.69/5.34 & 0.6055 & 0.4544 & 0.4562 & 2.66/9.17 \\
       +GPT2     & 0.6116 $\uparrow$ & 0.4549 & 0.4655 $\uparrow$ & 1.55/4.28 & 0.6065 $\uparrow$ & \textbf{0.4595} $\uparrow$ & \textbf{0.4655} $\uparrow$ & 2.74 $\uparrow$/9.82 $\uparrow$ \\ 
       \midrule
       $attention$ & 0.6128 & 0.4634 & 0.4700 & 1.93/6.09 & 0.5969 & 0.4491 & 0.4491 & 2.67/9.43 \\
       +GPT2     & 0.6142 $\uparrow$ & 0.4517 & 0.4755 $\uparrow$ & 0.67/1.94 & 0.6058 $\uparrow$ & 0.4536 $\uparrow$ & 0.4597 $\uparrow$ & 2.74 $\uparrow$/9.84 $\uparrow$ \\ 
       \midrule
       $max$       & 0.6092 & 0.4568 & 0.4611 & \textbf{2.49/8.72} & 0.6064 & 0.4574 & 0.4623 & 2.56/9.14 \\
       +GPT2     & 0.6138 $\uparrow$ & 0.4549 & 0.4655 $\uparrow$ & 0.67/1.81 & \textbf{0.6089} $\uparrow$ & 0.4503 & 0.4653 $\uparrow$ & 2.42/8.01 \\ 
       \midrule
       $mean$      & 0.6130 & 0.4633 & 0.4665 & 1.82/5.93 & 0.5998 & 0.4506 & 0.4534 & 2.65/9.60 \\
       +GPT2     & 0.6120 & 0.4557 & 0.4701 $\uparrow$ & 0.70/2.02 & 0.6063 $\uparrow$ & 0.4544 $\uparrow$ & 0.4609 $\uparrow$ & 2.74 $\uparrow$/9.80 $\uparrow$ \\ 
       \bottomrule
\end{tabular}
\end{table}

We can see from Table~\ref{pretrainU} and~\ref{pretrainO} that the performance of the proposed Static and Dynamic attention models can be improved by integrating the GPT2 representation on both Ubuntu and Opensubtitles datasets. 
It reveals that the pretrained GPT2 representation is better on capturing the semantics of the generated dialogue responses than the initial dialogue representation obtained by the RNN-based models.
In addition, the performance of five hybrid attention models can be improved by the integration of GPT2 representation, which is consistent to the performance of Static and Dynamic attention models. 
The results on both Ubuntu and Opensubtitles datasets are consistent, which actually reveals the improvement of the robustness of the proposed Static and Dynamic attention models. 
The reason may be that the integration of GPT2 representation could balance the models' performance between embedding metric and distinct metric on the two datasets and make the proposed models more stable in dialogue response generation. 

\begin{table}[htbp]
  \caption{The cases of generated responses by the proposed Static and Dynamic models with GPT2 on Ubuntu and Opensubtitles datasets.}\label{case-GPT2}
  \centering
    \begin{tabular}{l|l} 
      \multicolumn{2}{l}{\textbf{Ubuntu}}  \\
      \toprule
      \multicolumn{2}{l}{Case 1} \\ 
      \midrule
      \multirow{3}*{Context}& is there any way to install tomcat 5.5 on 12.04 ? \\
                            & wasn't it tomcat55 package \\ 
                            & apologies , it doesn't seem to be available anymore \\ 
                            \midrule
    Golden                   & tried tomcat 5.5 gives me a regex output and 5 isn't there \\ 
    \midrule
   Dynamic     & i	was	thinking	of	it	,	but	it	might	be	a	bug	in	launchpad  \\
   Dynamic + GPT2           & you can also search google for it and search for the package you need  \\
   Static    & i	think	it	's	not	a	package	that	is	called	"	tomcat	" \\
   Static + GPT2            & I'm not sure what 's the problem ,    \\ 
    \midrule
   \multicolumn{2}{l}{Case 2} \\ 
   \midrule
   \multirow{3}*{Context}& my dns is working good , I also cleared them . Delays still present \\
                            &  did you try disabling ipv6 with the boot option : ipv6 . \\
                            & trying , can we make the disable function working without restarting ? \\ 
                            \midrule
    Golden                        & may be able to use sysctl but not sure \\
    \midrule
    Dynamic    & I don’t know , but I don’t think it ’s possible to do that .  \\
    Dynamic + GPT2           & yes , I did that , but I don't know what to do .   \\
    Static    & I	don't	know	.	You	might	try	to	ping	the	gateway	address	.   \\ 
    Static + GPT2            & I'm not sure what you mean   \\ 
    \midrule    
    \multicolumn{2}{l}{\textbf{Opensubtitles}}  \\
    \midrule
    \multicolumn{2}{l}{Case 1} \\ 
    \midrule
    \multirow{3}*{Context}& What 's the matter with you ? \\
                            & You gotta be sitting ? \\
                            & Yeah , you 'll end up as a rug . \\ 
                            \midrule
   Golden                   & Come on , you 're holding up the works , you mug . \\ 
   \midrule
   Dynamic    & I	'll	take	care	of	you	.  \\
   Dynamic + GPT2           & I 'll get you out of the way .  \\
   Static    & You	're	a	fool	,	you	know	,	you	're	a	cinch	.   \\ 
   Static + GPT2            & You 're a fool .  \\ 
   \midrule
   \multicolumn{2}{l}{Case 2} \\ 
   \midrule
   \multirow{3}*{Context}& She 's a novel twist . \\
                            & Of course , I know you have no liking for Miss Hatton . \\ 
                            & She came in here , got nasty , I had to put her in her place . \\ 
                            \midrule
    Golden                   & Yes , I heard you had some slight misunderstanding \\ 
   \midrule
    Dynamic    & She	's	a	lot	.  \\
    Dynamic + GPT2           & I	'm	sorry	,	but	I	don	't	see	any	thing	to	see	you	,  \\
    Static    & Yes , And	I	'm	not	interested	in	her	either	.  \\ 
    Static + GPT2            & She was a bit of a high school girl like you didn 't like  \\ 
   \bottomrule
    \end{tabular}
\end{table}

Table~\ref{case-GPT2} shows the examples of generated responses by the proposed Static and Dynamic models with the integration of GPT2 on the Ubuntu and Opensubtitles datasets, respectively. 
From Table~\ref{case-GPT2} , we can see that the integration of GPT2 can improve the diversity of response generation. 
Furthermore, by integrating the pretrained GPT2, the proposed model may generate relevant but variant responses, for example "you can also search google for it and search for the package you need" (Dynamic + GPT2) is a relevant response, but is quite diverse comparing to the golden response.

\subsubsection{Human Evaluation}
For human evaluation, two evaluation metrics are chosen, namely \textbf{Coherence} and \textbf{Naturalness}.
As the example shown in Table~\ref{example-intro}, in multi turn dialogue generation, a generated response should not only dependent on the last input utterance, but should also consider the dialogue context in a session.
\textbf{Coherence} is used to measure the contextual relevance between a generated response and the context.
The Coherence score is one of 0,1 or 2, where 0, 1, 2 denote \emph{incoherent}, \emph{neutral} and \emph{coherent}, respectively.

However, in some cases, a contextual relevant response generated by a model may not be as natural as human responded.
For example, an input message, ``\emph{Can you tell me the way to the nearest bazaar?}'', the generated response ``\emph{Yes, I can tell you the way to the nearest bazaar.}'' is relevant to the input. 
But it is not a natural response.
Another example of an input and response pair is ``\emph{I don't know what you are talking about!}'' and ``\emph{I don't know what you are talking about!}''.
From the above examples, we can see that besides the Coherence, the naturalness of a generated response is also an important measure. 
Therefore, we proposed a metric to evaluate the naturalness of a generated response.
For human evaluation, given a dialogue context and a response generated by a model, \textbf{Naturalness} denotes whether the response can be an alternative to a human response.
The Naturalness score is 1 or 0, which indicates that the generated response is from a human or a model.

Besides the human evaluation metrics, we also automatically calculate the \textbf{Diversity} score of the generated responses by all the baselines and our model.
Here, for a generated response, we first calculate the number of distinct tokens in the response and then divide to the total number of distinct tokens in its dialogue context, which includes the number of distinct tokens in the given response.
Finally, the Diversity score equals to the arithmetic average of the above merchant of all the generated responses in test set.

For human evaluation, we randomly sample 500 test results of each model from Ubuntu and Opensubtitles test sets, respectively.
Three annotators, who are undergraduate students and not involved in other parts of the experiment, is asked to provide the evaluation scores. 
The final score of each model in test set equals to the arithmetic average of the three annotators .
Table~\ref{humaneva} shows the results of human evaluation on Ubuntu and Opensubtitles datasets.

\begin{table}[htbp]
\caption{The results of human evaluation on Ubuntu and Opensubtitles datasets. }
\label{humaneva}
\begin{tabular}{l|ccc|ccc}
\toprule
\multirow{2}{*}{\bf Models} & \multicolumn{3}{c|}{\bf Ubuntu} & \multicolumn{3}{c}{\bf Opensubtitles} \\
& \bf Coherence & \bf Naturalness & \bf Diversity & \bf Coherence & \bf Naturalness & \bf Diversity  \\
\midrule
LSTM & 0.930 & 0.477 & 0.069 & 0.963 & 0.443 & 0.099 \\
HRED & 0.967 & 0.490 & 0.141 & 0.963 & 0.443 & 0.098 \\
VHRED & 1.010 & 0.507 & 0.140 & 0.986 & 0.473 & 0.093 \\
CVAE & 0.987 & \bf 0.513 & 0.140 & 1.000 & 0.477 & \bf 0.114 \\
WSI & 1.010 & 0.507 & 0.141 & 1.013 & 0.490 & 0.110 \\
HRAN & 1.027 & 0.510 & 0.147 & \bf 1.033 & 0.477 & 0.109 \\
ReCoSa & 1.050 & 0.490 & 0.148 & 1.010 &  \bf 0.540 & 0.110 \\
\midrule
Dynamic & 0.987 & 0.507 & \bf 0.158 & 1.013 & 0.477 & 0.109 \\
Static & \bf 1.070 & \bf 0.513 & 0.150 & 1.027 & 0.497 & 0.110 \\
\bottomrule
\end{tabular}
\end{table}

From table~\ref{humaneva}, we can see that the proposed static attention-based model outperforms the baselines in all the three evaluation metrics on Ubuntu dataset. 
For the Diversity, we can see that the proposed dynamic attention-based model is better than other baselines on Ubuntu dataset.
We also notice that the CVAE model obtains the best diversity performance and the best Naturalness performance is from ReCoSa model on Opensubtitles dataset.
Meanwhile, we also notice that the subjective experience of human to the generated responses (Coherence and Naturalness) is not positively correlated to the statistical evaluation metric, Diversity.  
It demonstrates that in human conversation, the way of presenting (the coherent and natural responses) may be more important than the content they saying (the use of diverse tokens). 

\subsubsection{$\bullet$~~Results on Various Context Lengths}
\begin{figure}[!ht]
\centering
\includegraphics[width=\linewidth]{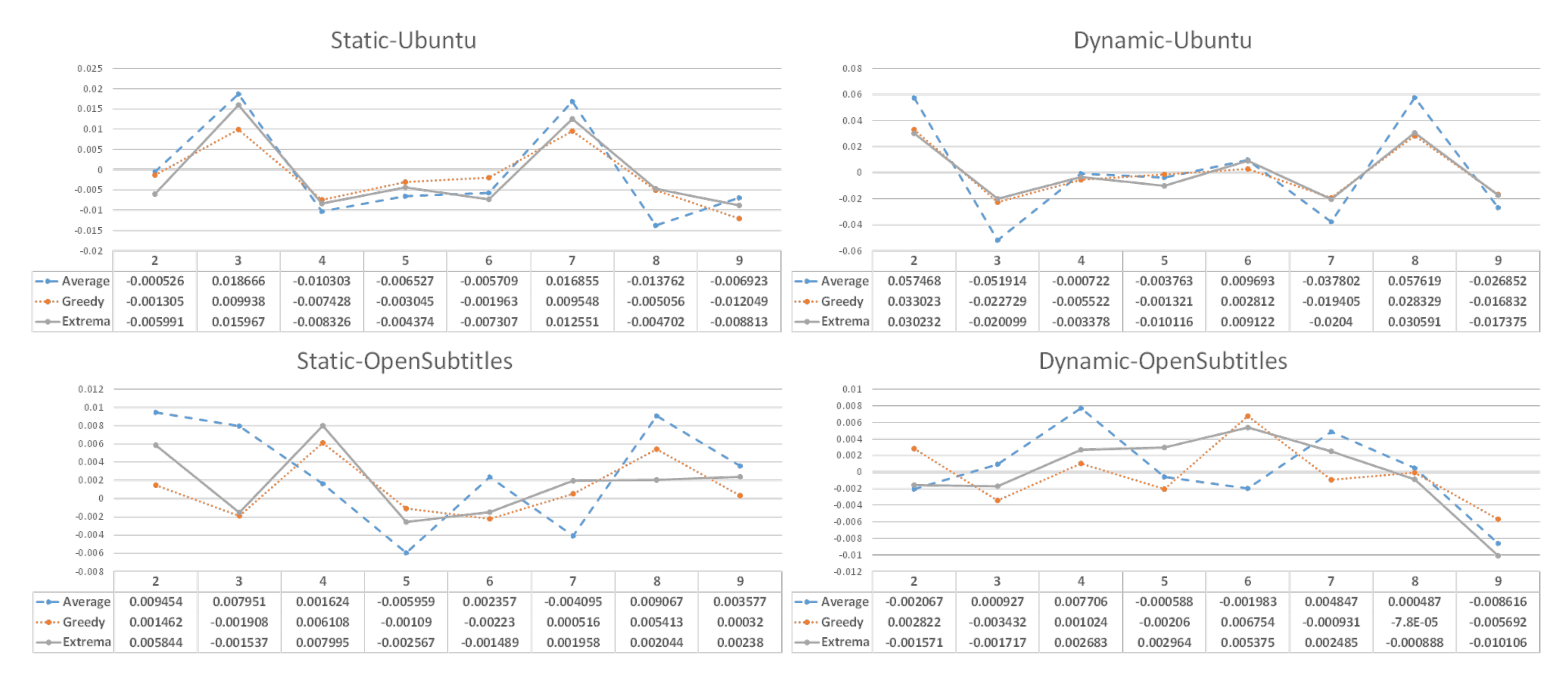}
\caption{The impact of context length on the performance of the Static and Dynamic models on Ubuntu and Opensubtitles datasets. The range of context length is from 2 to 9.}
\label{static-dynamic}
\end{figure}

To verify the impact of context length on the performance of the proposed model for response generation, we vary the length of context to respectively train the proposed models, which are called various context models, on two datasets.
Here, context length indicates the number of historical utterances that are used for encoding in a dialogue context.
Figure~\ref{static-dynamic} shows the results of the proposed static and dynamic attention models in various context length, respectively.

%====================================

The values denote the difference between the results of Static and Dynamic with varying of context length.
It also verifies that the generation of multi turn dialogues is a context-sensitive process, which is relevant to the encoded dialogue history.
We can draw the following conclusions through the observing and analysis of the results in Figure~\ref{static-dynamic}. 
\begin{itemize}
\item To compare the results on Ubuntu and Opensubtitles, we can see that the Ubuntu results in three embedding metrics are more stable and consistent than Opensubtitles results. 
The reason may be that the tokens used in the utterances of Ubuntu dataset are more formal than those in Opensubtitles dataset. 
Taking the cases in Table~\ref{case-GPT2} as example, we can see the utterances in Opensubtitles dataset are shorter and more casual than those in Ubuntu dataset. 
Meanwhile, the number of entities in the utterances of Ubuntu dataset are more than those of Opensubtitles dataset.
Therefore, the variance of results in Opensubtitles dataset is larger than in Ubuntu dataset. 
\item  To compare the results of Static and Dynamic models,  we can find that their performance trend is different in the various context length.  
To see the results on Ubuntu dataset, we can sum up the reason of the experimental results that the capability of Static and Dynamic models on utilizing the context is different. 
For the Dynamic model, it needs a greater number of utterances to dynamically select the useful information to encoding and then generate a proper response. 
However, for the Static model, a fix number of utterances, which usually the last three utterances in the Ubuntu dataset, may be enough for response generation.  
\end{itemize}

\subsubsection{Additional Analysis with SOTA model}
\begin{figure}[!ht]
\centering
\includegraphics[width=\linewidth]{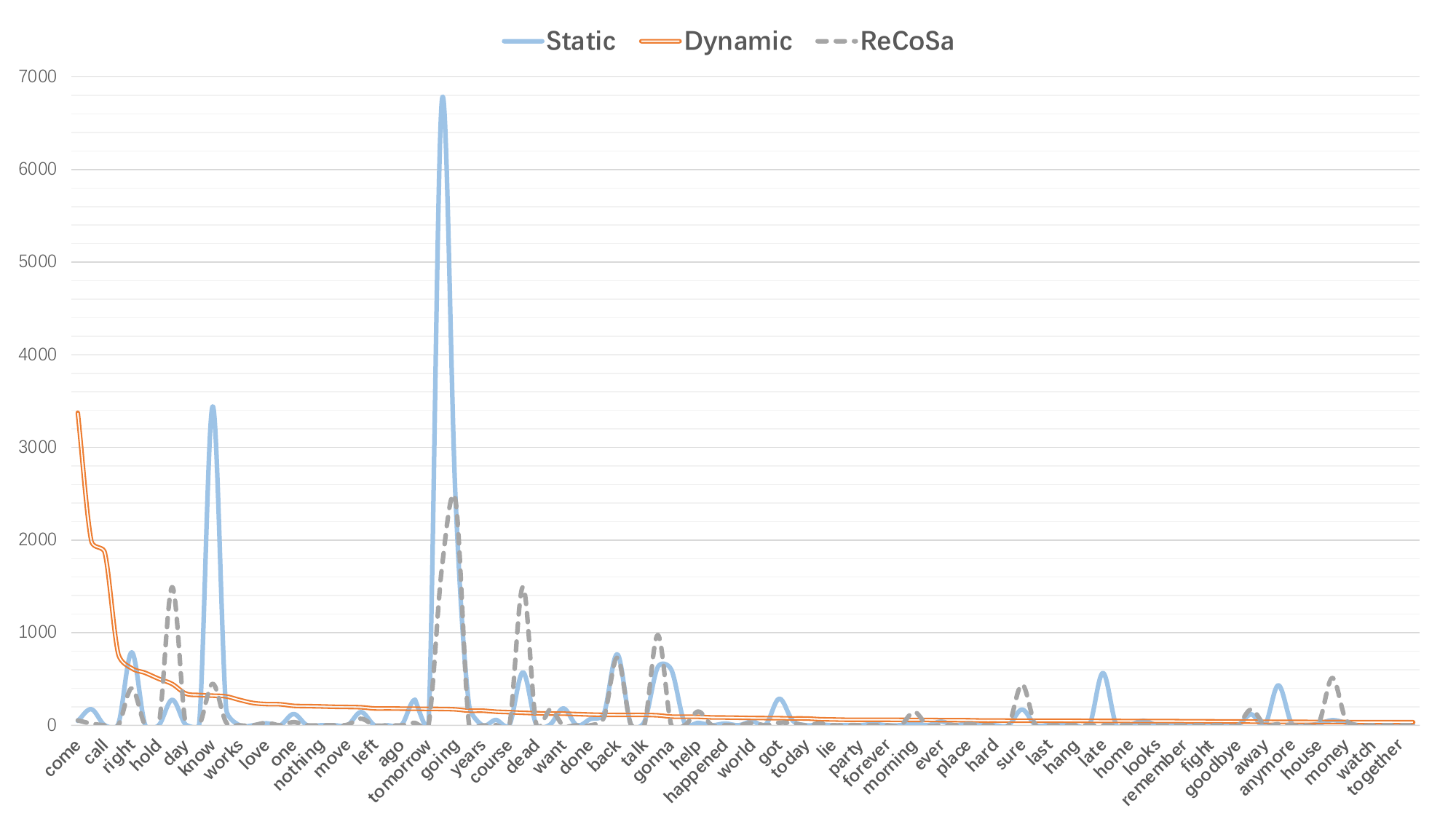}
\caption{The statistics of the token frequencies of the static, dynamic attention-based models and the ReCoSa model on the vocabulary. }
\label{tfdis}
\end{figure}

For the analysis of the generated cases of the proposed models and the SOTA baseline (ReCoSa) in test set, we first present some token-level statistics to show the difference of the models in the generation of multi turn dialogues. 
We calculate the frequency distributions of the tokens of the generated dialogue responses by the three models over the vocabulary. 
Note that the stop words are removed from the vocabulary. 
Figure~\ref{tfdis}  shows the statistics of the token frequencies of the proposed static and dynamic attention-based models and the ReCoSa model. 

From Figure~\ref{tfdis}, we can see that the token frequency distributions of the three models are quite different. 
It demonstrates that given the same set of input messages, the generate responses are various. 
The reasons may be two-fold:
\begin{itemize}
\item First, the most essential characteristic of the open domain dialogue generation is the one-to-many phenomenon, which denotes that given an input message, there may be many suitable and reasonable responses. 
Therefore, the dialogue model needs to learn the one-to-many relations and the generation of the dialogue responses may be various among different models.  
\item Second, the attention mechanisms in the three models are different so that the distributions of token weights in the vocabulary may have large discrepancies in decoding phase. 
\end{itemize}

\begin{figure}[!ht]
\centering
\includegraphics[width=0.85\linewidth]{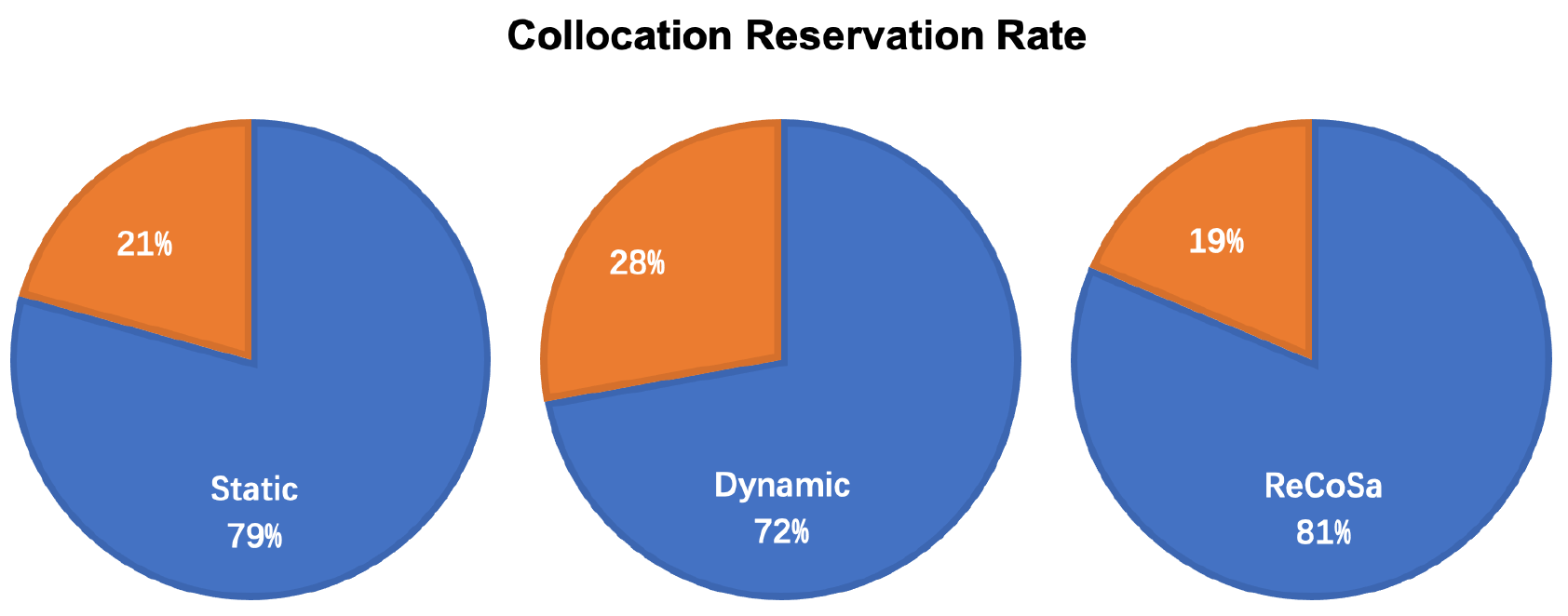}
\caption{The collocation reservation rate of the static, dynamic attention-based models and the ReCoSa model on test set. }
\label{crr}
\end{figure}

Furthermore, we also calculate the collocation reservation rate of the static, dynamic attention-based models and the ReCoSa model. 
Here, a collocation indicates a token pair, where the first token and the second token come from the input message and response respectively. 
For a dialogue model, the collocation reservation rate thus equals to the number of collocations generated by the model divides by the number of collocations in the test set. 
Figure~\ref{crr} presents the collocation reservation rate of the three models. 

Here, as shown in Figure~\ref{crr}, the reservation rate can be partially seen as a capability of fitting the gold responses. 
However, we should also notice that the high collocation reservation rate may impact the performance of diversity of the generation of dialogue responses. 

\begin{figure}[!ht]
\centering
\subfigure[Static]{
\centering
\includegraphics[width=0.45\linewidth]{fig/static_wcloud.pdf}
}
\subfigure[Dynamic]{
\centering
\includegraphics[width=0.45\linewidth]{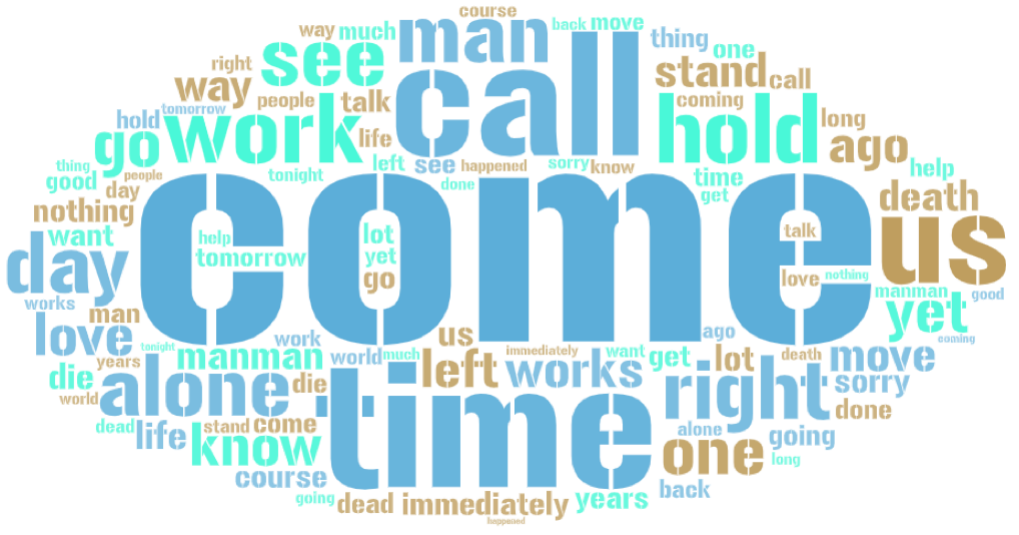}
}
\subfigure[ReCoSa]{
\centering
\includegraphics[width=0.45\linewidth]{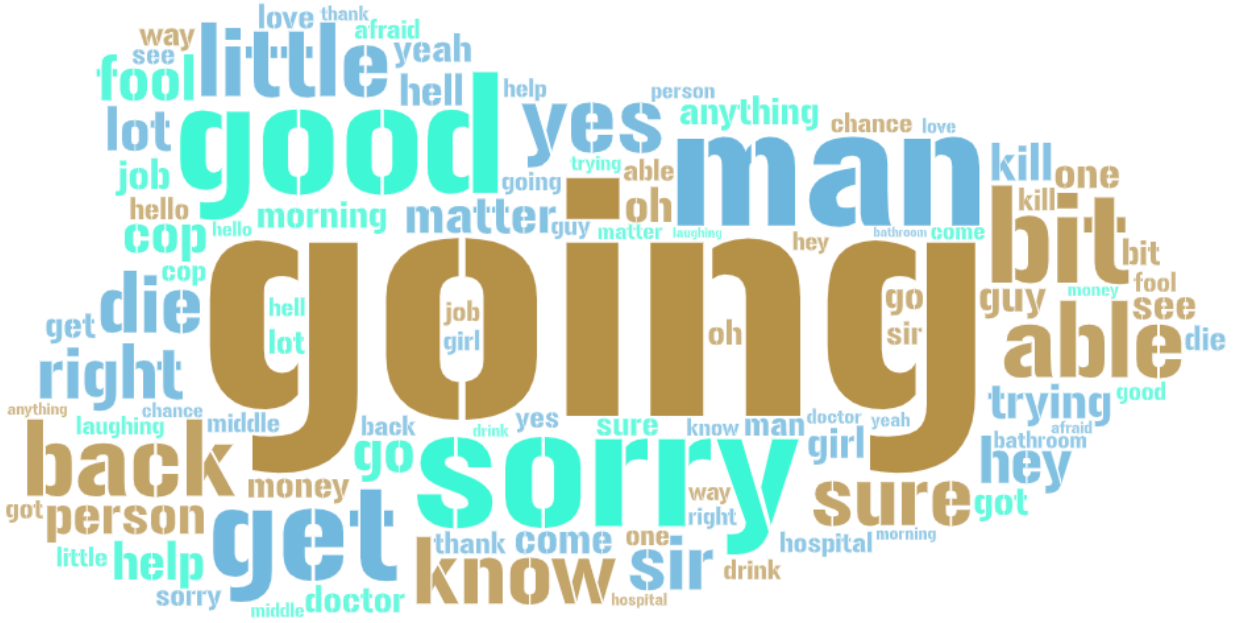}
}
\caption{The word clouds of the generated responses by the proposed static (a), dynamic (b) attention-based models and the ReCoSa (c) model on the test set.}
\label{wcloud}
\end{figure}

To exhibit the frequently generated tokens by the three models, we also draw the word clouds as shown in Figure~\ref{wcloud}. 

%\newpage

\section{Conclusion and Future Work}

In this paper, we proposed a static and dynamic attention-based approach to the generation of open domain multi turn dialogues. 
The proposed static and dynamic attention-based model is an attention-based framework for multi-turn dialogue generation. 
We have verified that it is compatible to different granularities of input information, such as token-level and utterance-level information, different types of attentions, such as multi head attention and self-attention, and pretrained language model, such as GPT2, etc. 
Furthermore, the proposed static and dynamic attentions can be integrated in various measures, such as concatenation, direct sum, weight sum (through learnable parameters and attention), max and mean pooling. 
Experimental results showed that the proposed framework is generally better than the baselines in automatic and human evaluations.
Various aspects of the proposed model have beeb analyzed through the case studies.  
Finally, we also analyzed some factors that impact the performance of the proposed models and the SOTA model. 
The token distribution, the collocation reservation rate and the context length on the performance of the proposed models for multi turn dialogue generation.
In future work, we plan to consider the linguistic phenomenons, such as ellipsis and anaphor, in modeling the dialogue context and generate more coherent responses. 

\section{Acknowledgments}
%
%Identification of funding sources and other support, and thanks to
%individuals and groups that assisted in the research and the
%preparation of the work should be included in an acknowledgment
%section, which is placed just before the reference section in your
%document.

This paper is supported by the Science and Technology Innovation 2030 Major Project of China (No. 2020AAA0108605) , National Natural Science Foundation of China (No. 62076081, No. 61772153 and No. 61936010) and Nature Scientific Foundation of Heilongjiang Province (No.YQ2021F006).

%\newpage
\bibliographystyle{ACM-Reference-Format}
\bibliography{sample-base}

\end{document}